\documentclass[sigconf, anonymous = False]{acmart}
\AtBeginDocument{%
  \providecommand\BibTeX{{%
    \normalfont B\kern-0.5em{\scshape i\kern-0.25em b}\kern-0.8em\TeX}}}

\setcopyright{acmcopyright}
\copyrightyear{2018}
\acmYear{2018}
\acmDOI{10.1145/1122445.1122456}

\acmConference[Woodstock '18]{Woodstock '18: ACM Symposium on Neural
  Gaze Detection}{June 03--05, 2018}{Woodstock, NY}
\acmBooktitle{Woodstock '18: ACM Symposium on Neural Gaze Detection,
  June 03--05, 2018, Woodstock, NY}
\acmPrice{15.00}
\acmISBN{978-1-4503-XXXX-X/18/06}

\usepackage{color}
\usepackage{makecell}
\usepackage{multirow}
\usepackage{graphicx}
\usepackage{subfigure}
\usepackage{svg}
\usepackage{sidecap}
\begin{document}
\settopmatter{printacmref=false} % Removes citation information below abstract
\renewcommand\footnotetextcopyrightpermission[1]{} % removes footnote with conference information in first column
\pagestyle{plain} % removes running headers

\title{Multi-Teacher Knowledge Distillation for \\Incremental Implicitly-Refined Classification}


 \author{ Longhui Yu}
 \email{ yulonghui@stu.pku.edu.cn}
 \affiliation{%
  \institution{Peking University}
  \country{China}}

\author{Zhenyu Weng}
 \email{wzytumbler@pku.edu.cn}
\affiliation{%
  \institution{Peking University}
  \country{China}}

\author{Yuqing Wang}
 \email{wyq@stu.pku.edu.cn}
\affiliation{%
  \institution{Peking University}
  \country{China}}
 
 \author{Yuesheng Zhu}
  \email{zhuys@pku.edu.cn}
\affiliation{%
  \institution{Peking University}
  \country{China}}

\renewcommand{\shortauthors}{Trovato and Tobin, et al.}

\begin{abstract}

% As an extension to the class incremental learning, Incremental Implicitly-Refined Classification  (IIRC) aims at simulating human cognition to continually learn new classes, discover their granularity (subclass), and still preserve previous knowledge (superclass). However, distillation-based class incremental learning methods sequentially distill knowledge only from the last model, which cannot work for IIRC because the previous superclass knowledge may be occupied by subclass knowledge in the incremental learning process. To solve this problem, we propose a novel Multi-Teacher Knowledge Distillation (MTKD) strategy, in which a student model is trained by using the knowledge from multiple teacher models for superclasses and subclasses. To preserve the subclass knowledge, we use the last model as a general teacher to distill previous knowledge for the student model. To preserve the superclass knowledge, we use the initial model as a superclass teacher to distill the superclass knowledge as the initial model contains abundant superclass knowledge. In addition, we propose a general post-processing mechanism, called as Top-k prediction restriction. This mechanism can combine with the existing incremental learning methods to solve the redundant prediction problem in IIRC. Our experimental results on IIRC-ImageNet120 and IIRC-CIFAR100 show that the proposed method can achieve better classification accuracy compared with existing state-of-the-art methods.

Incremental learning methods can learn new classes continually by distilling knowledge from the last model  (as a teacher model) to the current model  (as a student model) in the sequentially learning process. However, these methods cannot work for Incremental Implicitly-Refined Classification  (IIRC), an incremental learning extension where the incoming classes could have two granularity levels, a superclass label and a subclass label. This is because the previously learned superclass knowledge may be occupied by the subclass knowledge learned sequentially. To solve this problem, we propose a novel Multi-Teacher Knowledge Distillation  (MTKD) strategy. To preserve the subclass knowledge, we use the last model as a general teacher to distill the previous knowledge for the student model. To preserve the superclass knowledge, we use the initial model as a superclass teacher to distill the superclass knowledge as the initial model contains abundant superclass knowledge. However, distilling knowledge from two teacher models could result in the student model making some redundant predictions. We further propose a post-processing mechanism, called as Top-k prediction restriction to reduce the redundant predictions. Our experimental results on IIRC-ImageNet120 and IIRC-CIFAR100 show that the proposed method can achieve better classification accuracy compared with existing state-of-the-art methods.

% To further solve the redundant prediction problem in IIRC, we propose a general post-processing mechanism, called as Top-k prediction restriction. This mechanism can also combine with the existing incremental learning methods. Our experimental results on IIRC-ImageNet120 and IIRC-CIFAR100 show that the proposed method can achieve better classification accuracy compared with existing state-of-the-art methods.
%改改改
\end{abstract}




%%
%% Keywords. The author (s) should pick words that accurately describe

\maketitle

\begin{figure}[ht]
\centering
\includegraphics[scale=0.45]{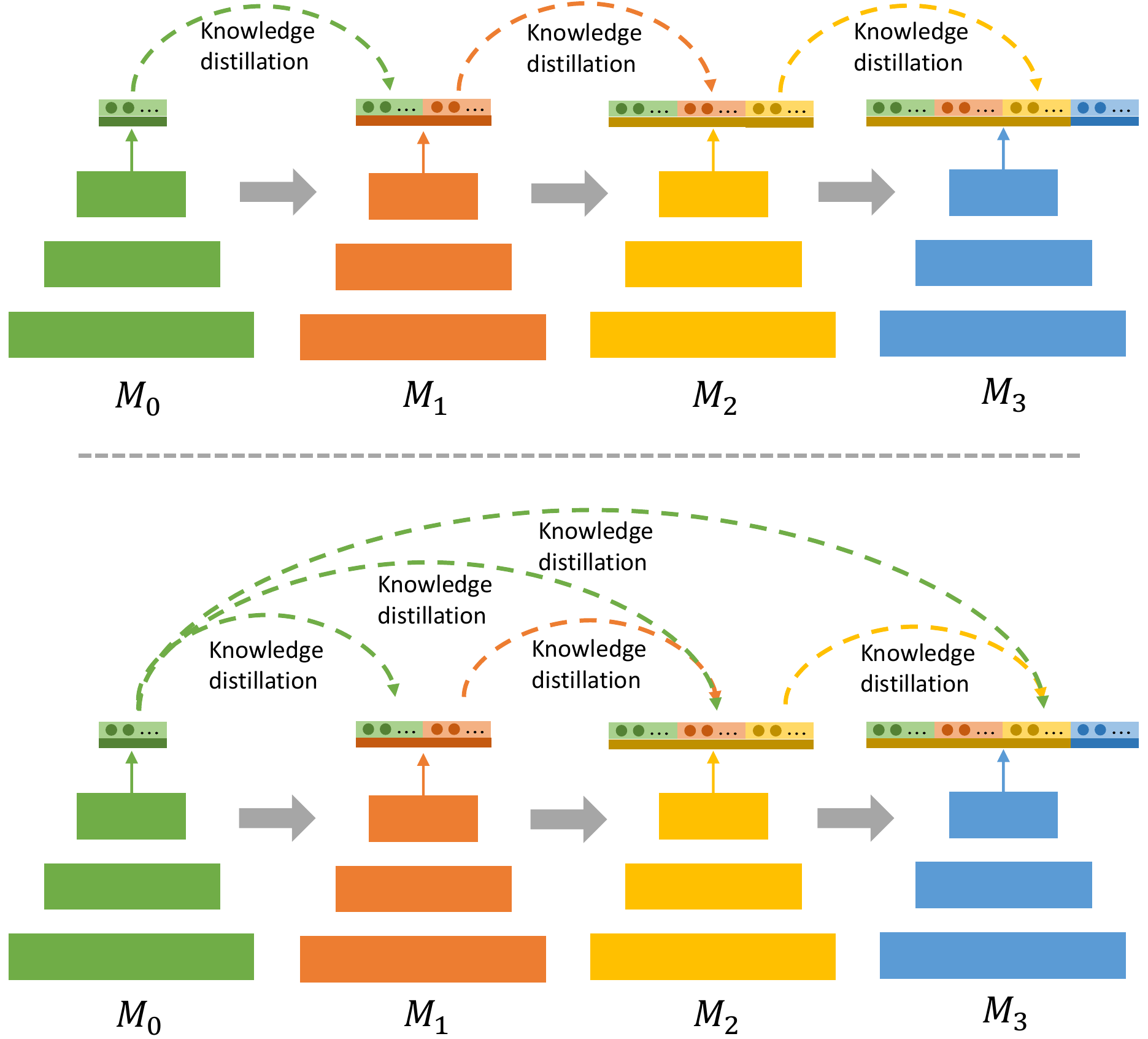}
\caption{Comparison between conventional incremental learning methods  (top) and our MTKD strategy  (down). Conventional incremental learning methods distill knowledge sequentially. To consolidate the knowledge learned in the initial model which contains great superclass knowledge, the student model in our MTKD strategy learns knowledge from both the initial model  (superclass teacher) and the last model  (general teacher).}
\label{fig1}

\end{figure}
\section{Introduction}

In recent years, deep learning methods have made tremendous progress in image classification~\cite{he2016deep, simonyan2014very, krizhevsky2012imagenet}, speech recognition~\cite{abdel2014convolutional, hinton2012deep}, video recommendation~\cite{covington2016deep, zhang2019deep} and other multimedia applications. Although deep learning methods can either match or surpass human performance in some applications, such progress is limited to a narrow setting where the number of training classes is fixed. Humans can continually learn and accumulate knowledge. In contrast, the current learning methods learn from the new tasks would have a significant performance drop on the old tasks, known as catastrophic forgetting~\cite{mccloskey1989catastrophic}.

To simulate an incremental visual system that continually uses new classes to extend the model's knowledge, increment learning methods~\cite{hu2021distilling, rebuffi2017icarl, li2017learning} use knowledge distillation~\cite{hinton2015distilling} to continually train a model to learn new classes while preserving the knowledge learned from old classes. Specifically, they distill knowledge from the last model  (as a teacher model) to the current model  (as a student model) in the sequentially learning process. However, the incremental learning setting is not always satisfied in real-world scenarios since humans not only know new entities but also discover new information about them after interacting with them multiple times. Hence, Incremental Implicitly-Refined Classification  (IIRC)~\cite{abdelsalam2020iirc} is designed, which is an extension to the incremental learning setup and more aligned with the real-world scenario. In IIRC, classes follow a hierarchical relationship  (e.g., Polar Bear is a subclass of Bear). Instead of predicting just one label of each sample, the model is required to learn superclass at first and find the relationship between superclass and subclass when learning subclass that appears later.

\begin{figure}[t]
\centering
\includegraphics[scale=0.25]{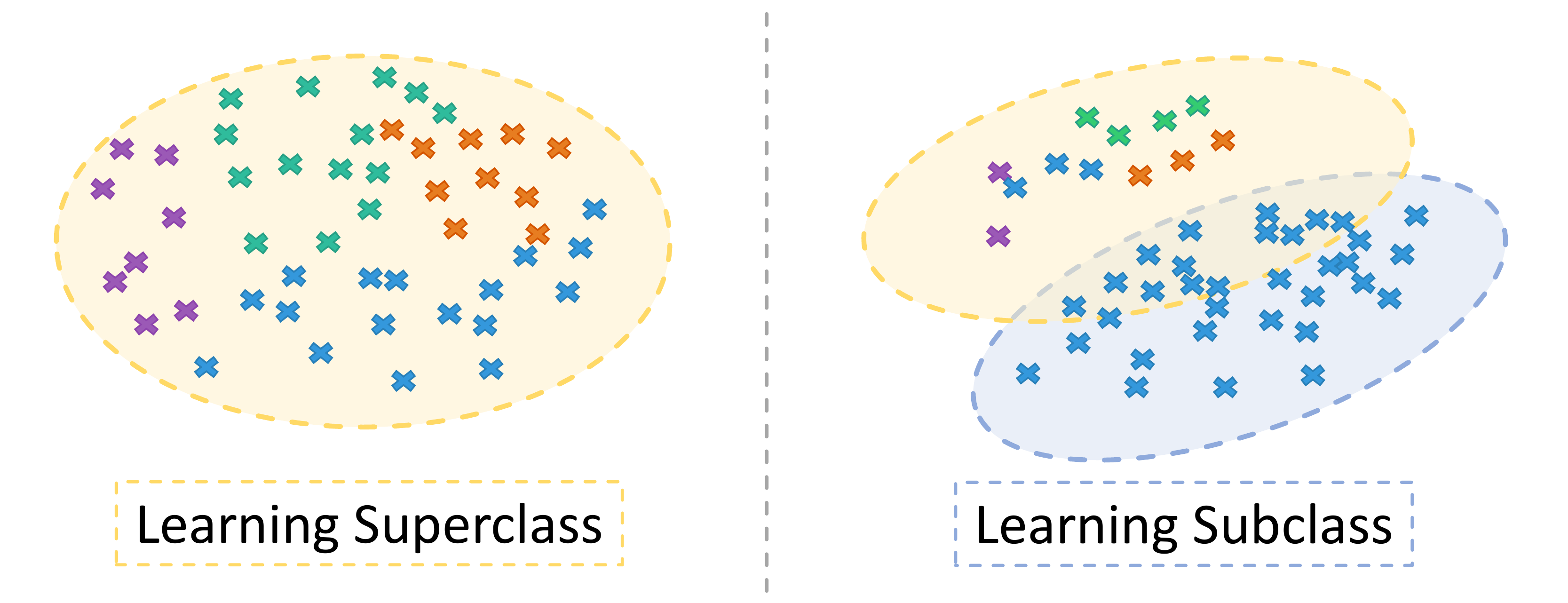}
\caption{Comparison between learning superclass and learning subclass. When model learns superclass, a plenty of superclass labeled samples could support the model to learn an unbiased decision boundary  (left). While the model learns a specific subclass, serious data imbalance between superclass and subclass could cause that superclass knowledge is occupied by subclass knowledge  (right).}
\label{squeeze}

\end{figure}
%%改改改改改
Existing methods perform well on conventional incremental learning, however, they suffer from significant performance degradation in IIRC. We argue that this is due to the previous superclass knowledge may be occupied by subclass knowledge in the incremental learning process.

% b). A sigmoid-based incremental learning model would
% output some false positive predictions.

Concretely speaking, in IIRC, subclass samples are similar or even identical to that of the superclass. Existing work~\cite{deng2009imagenet, castro2018end} shows that, in incremental learning, when old samples and new samples come from similar tasks, learning new tasks could cause serious forgetting about old tasks. This kind of forgetting is aggravated by the data similarity between superclass and subclass in IIRC. As shown in Figure \ref{squeeze}, when the model needs to learn a new subclass task, since superclass and subclass have similar or even identical samples, this would result in the decision boundaries of the superclass and subclass being very close to each other. However, in IIRC, when learning subclass, the superclass label is not provided. A large number of subclass labeled samples could force the model to squeeze the decision boundaries of the superclass. In this way, the superclass knowledge would be occupied by the subclass knowledge effortlessly. 

% In terms of b), in IIRC, the model is required to output one or two predictions. Building a binary classifier with a sigmoid activation function is a straightforward way for multi-label classification task~\cite{chen2019multi}. However, a sigmoid classifier for multi-label classification is prone to output too many wrong predictions~\cite{chen2019multi}.

%  where a learner usually learn a series of entities from the same family in multiple interactions, discovers more granularity about them, while still trying not to forget previous knowledge

% \begin{figure}[ht]
% \centering
% \includegraphics[scale=0.25]{fig2.png}
% \caption{The blue bar shows Jaccard similarity of different incremental learning method in a 10 steps incremental setting on CIFAR100, while the orange bar shows the Jaccard similarity in a 11 steps IIRC setting. The same method has a significant performance drop in IIRC.}
% \label{fig:pathdemo4}
% \end{figure}

To solve the issue, we propose a Multi-Teacher Knowledge Distillation  (MTKD) strategy in this paper. In our proposed strategy, there are two teacher models applied to train a student model. Following the conventional incremental learning methods~\cite{li2017learning, douillard2020podnet, rebuffi2017icarl, hou2019learning, wu2019large}, one teacher model is the model preserved in the last incremental step, which is called general teacher. The general teacher is applied to teach the student model the knowledge of classes appearing subsequently, except for superclass appearing in the initial step. To consolidate superclass knowledge, in this work, we use the model in the initial incremental step as another superclass teacher to teach the student model defective knowledge of superclass. Figure \ref{fig1} shows the difference between conventional incremental learning methods and our MTKD strategy. However, distilling knowledge from two teacher models could result in the student model making some redundant predictions. To further solve the redundant prediction problem, we design a simple Top-k prediction restriction mechanism. K represents the max activated prediction number per image and is decided by the hierarchical label structure. In this paper, our contributions are as follows: 
\begin{itemize}
  \item We propose a Multi-Teacher Knowledge Distillation strategy for IIRC to prevent superclass knowledge from being occupied by the knowledge of subclass.
  \item To further tackle the redundant prediction problem caused by using multiple teacher models in MTKD, we propose a simple yet effective Top-k prediction mechanism combined with our MTKD strategy to reduce the unnecessary predictions.
  \item Detailed analysis shows that superclass knowledge could be occupied by the subclass knowledge in IIRC and our MTKD strategy can maintain superclass discrimination effectively.
  \item Experiments show that the proposed Top-k prediction restriction mechanism can combine with our MTKD strategy (k-MTKD) conveniently and outperform existing state-of-the-art by a large margin. The proposed k-MTKD strategy gains $11.5\%$ improvement on IIRC-CIFAR100 and $25.7\%$ on IIRC-ImageNet120.
\end{itemize}

%改改改改改

%  (1) We show existing methods perform on IIRC poorly, since superclasses knowledge may be replaced by their subclasses;  (2) We propose a Multi-Teacher Knowledge Distillation strategy to improve the accuracy by maintaining the discrimination of superclass;  (3) To conquer the redundant predictions problem caused by our MTKD strategy, we propose a simple Top-k prediction restriction mechanism to reduce redundant prediction;  (4) Experiments show the MTKD strategy can combine with existing methods conveniently and improve the classification performance.

\section{Related Work}

Enabling machines to imitate human cognition is an inspirational task. Incremental learning as a learning paradigm, which can save training costs and storage space consumption, has been an active topic for a long time~\cite{cauwenberghs2001incremental, ruping2001incremental}. Existing incremental learning work can be divided into $parameter-based$ and $distillation-based$ methods by the way they tackle the problem of catastrophic forgetting.

\textbf{Parameter-based.} The basic idea of parameter-based methods is to reduce the changes between the old model's parameters and new model's when learning new tasks. Some works like EWC~\cite{kirkpatrick2017overcoming}, SI~\cite{zenke2017continual}, RWalk~\cite{chaudhry2018riemannian}, and MAS~\cite{aljundi2018memory} try to find important parameters for both old tasks and new tasks. They find these important weights by different methods and penalize the change on these parameters. 

\textbf{Distillation-based.}  Instead of penalizing parameters of models across different tasks, distillation-based incremental learning methods~\cite{castro2018end,douillard2020podnet, li2017learning,zhang2020class,javed2018revisiting,rebuffi2017icarl,hou2019learning,dhar2019learning} try to align logtis, features, attention maps between the old model and the new model. Existing distillation-based incremental learning methods usually combine with rehearsal strategy to defy catastrophic forgetting. icarl~\cite{rebuffi2017icarl} is a typical baseline in this field, they use a herding~\cite{welling2009herding} strategy to select representative examples for old knowledge preserve. Besides, Training networks with Knowledge distillation loss and a nearest-mean-of-exemplars classifier make icarl perform well on both old tasks and new tasks. E2E~\cite{castro2018end} propose a class-balanced finetuning strategy to alleviate the preference for new tasks. LUCIR~\cite{hou2019learning} distills feature knowledge from the old model and proposes a cosine normalization classifier to replace the standard softmax layer which has a prominent preference for new samples. PODNet~\cite{douillard2020podnet} proposes a novel attention-based feature knowledge distillation loss to better preserve old knowledge, and uses a multiple proxy classifier to alleviate the influence produced by the change of feature extraction network. In this paper, we focus on the distillation-based methods that are applied in IIRC. In contrast to existing distillation-based incremental learning methods, our proposed MTKD strategy distill logits knowledge from two teachers. One is for general knowledge distillation; the other is for distilling coarse superclass knowledge.

\begin{figure*}[ht]

\centering
\includegraphics[scale=0.48]{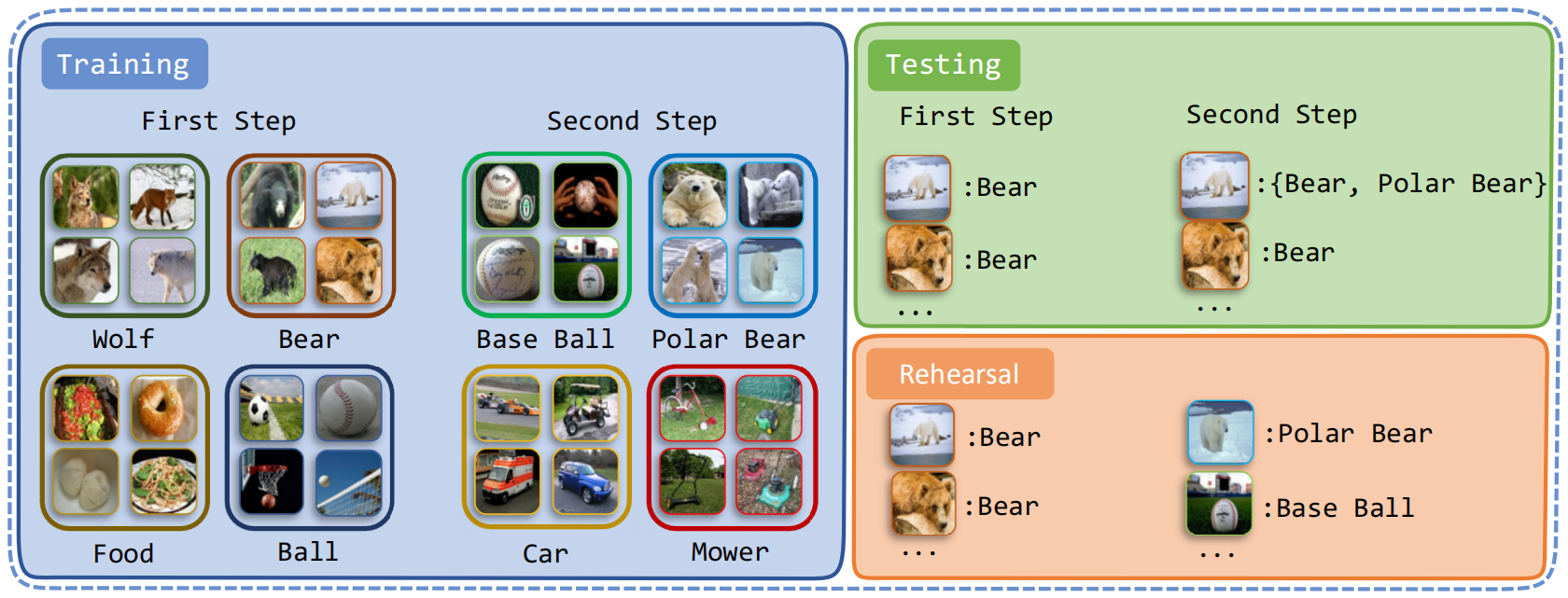}
\caption{Training, testing, and rehearsal process in IIRC. In the training phase, superclasses appear early than their subclasses, and only one label is provided for training. In the testing phase, the model is required to predict all the labels it has seen. As for the rehearsal strategy, the preserved label is that used in the training phase, and only one label is preserved.}
\label{IIRCYS}
\end{figure*}

\section{Background}
\subsection{Incremental Learning with Knowledge Distillation}

Before introducing our method, in this section, we describe the baseline of the rehearsal-based method using knowledge distillation in conventional class incremental learning. 

Assuming there are a total of $T+1$ increment steps, an initial incremental step and T subsequent steps. In the $t_{th}$ step, $0 \leq t \leq
T$. In the $t_{th}$ step, model would get some training data of several new tasks $\{{X^t, Y^t}\}$. $\{X^t = \{x_i^t\}_{i=1}^{N_t}\}$ and \{$Y^t = \{y_i^t\}_{i=1}^{N_t}, y_i^t \in [n+1, ..., n+m]\}$ are the training images of new tasks and ground-truth labels, where $N_t$ denotes the number of training samples in new task t, and n, m denote the number of previous classes and new classes. In all the incremental steps except the initial step, the select old exemplars are denoted as $\{{\hat{X}^t, \hat{Y}^t}\}$. $\{\hat{X}^t = \{\hat{x}_i^t\}_{i=1}^{N_o}\}$ and \{$\hat{Y}^t = \{\hat{y}_i^t\}_{i=1}^{N_o},  \hat{y}_i^t \in [1, ..., n]\}$ are the preserved images of previous tasks and ground-truth labels. where $N_o$ denotes the number of preserved samples before task t, and n denotes the number of previous classes. Hence, in the $t_{th}$ step, we have the training set composed of new samples and rehearsal samples, $D = X^t \cup \hat{X}^t$.

As for the training process, two loss functions, cross-entropy loss $L_{ce}$ and knowledge distillation loss $L_{kd}$ are used for learning new concept and meanwhile preserving previous knowledge. 

As for the softmax-based cross-entropy loss, it is computed as follow: 
\begin{equation}
    L_{ce} = \sum_{ (x,y) \in D} \sum_{i = 1}^{n+m} -y_i log[p_i (x)],
\end{equation}
where $y_i$ denotes whether the training data belongs to label $i$, $p_i (x)$ is the logits for class $i$ after a softmax normalization.

Let us denote the output logits of the old and new models as $\hat{O}^n (x) = [\hat{o}_1 (x), ..., \hat{o}_n (x)]$ and ${O}^{n+m} (x) = [{o}_1 (x), ..., {o}_n (x), {o}_{n+1} (x)$ $, ..., {o}_{n+m} (x)]$. The distillation loss is formulated as below:
\begin{equation}
    L_{kd} = \sum_{ (x,y) \in D} \sum_{i = 1}^{n} -\hat{\pi}_i (x) log[\pi_i (x)] ,
\end{equation}
\begin{equation}
    \hat{\pi}_i (x) = \frac{e^{\hat{o}_i (x)/ T}}{\sum_{j=1}^n e^{\hat{o}_j (x)/ T}}, 
    ~ {\pi}_i (x) = \frac{e^{{o}_i (x)/ T}}{\sum_{j=1}^n e^{{o}_j (x)/ T}},
\end{equation}   
where $T$ is the temperature scalar, which is commonly used in knowledge distillation to smooth the logit.

\begin{figure}[t]
\centering
\includegraphics[scale=0.65]{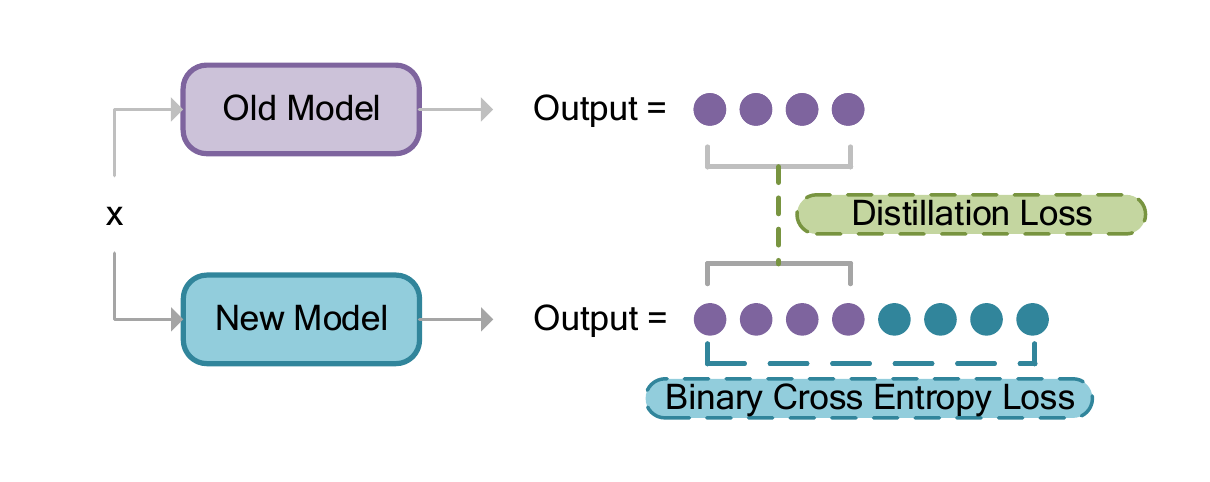}
\caption{Baseline used in this work, distillation loss is conducted on logits of previous classes between the old model and new model, while binary cross-entropy loss is computed on the output of all the classes. }
\label{common2}
\end{figure}

% \subsection{Incremental Implicitly-Refined Classification}
\subsection{Benchmark for IIRC}
% In IIRC~\cite{abdelsalam2020iirc}, two popular benchmark, CIFAR-100~\cite{krizhevsky2009learning}, ImageNet-1000~\cite{deng2009imagenet} are used to compose the IIRC benchmark. 

IIRC constructs a two-level hierarchy label structure. For each sample, it has one or two labels, including a superclass label and a subclass label. Notice there are also some classes that do not have parents. For each pair of superclass label and subclass label, the superclass knowledge is always introduced earlier than the subclass. In addition, the number of samples for the superclass is always more than that of its subclass. 

% , while in the IIRC-ImageNet1000, the number is 3 to 118. The superclass samples are selected from samples of its subclass. In the IIRC setup, they select 40\% subclass samples to their superclass and remain 80\% samples for subclass itself. This means there are 20\% overlapping between superclasses and subclasses. In IIRC-ImageNet1000, for some superclass which have more than 8 subclasses, uses $\frac { 8 } { number\  of\  subclasses}$ $\times$ $40\%$ of samples from its subclasses. 

The training, testing processing, and rehearsal strategy are shown in Figure \ref{IIRCYS}. In IIRC, the model is required to predict all the correct labels while testing. However, while the model learns subclass knowledge, although a sample could have a superclass label and a subclass label, only the subclass label is available to update the model. As for the rehearsal strategy, IIRC selects 20 examples per class for all the classes, including superclasses and subclasses. In IIRC, the label for the preserved sample is the training label when the sample acts as new data and would not change  (e.g., a Polar Bear image labeled Bear is preserved as a Bear image for all the subsequent steps).

\subsection{Baseline for IIRC}
The conventional incremental learning methods are not suitable for Incremental Implicitly-Refined Classification  (IIRC), as IIRC is a multi-label classification task. The model needs to output a superclass label and a subclass label if necessary. Hence, in IIRC~\cite{abdelsalam2020iirc}, the last softmax layer of the conventional incremental learning methods is replaced by a binary sigmoid layer for each class. Classes with the output value after sigmoid function above 0.5 are considered as activated predictions in inference. In IIRC, the binary cross-entropy loss is as follow:
\begin{equation}
    L_{bce} = \sum_{ (x,y) \in D} \sum_{i = 1}^{n+m} -y_i log[p_i (x)] +  (1-y_i)log[ (1 - p_i (x))].
\end{equation}

For the distillation loss, different from conventional incremental learning, there is a sigmoid activation after the last $FC$ layer instead of softmax activation. In IIRC, the distillation loss is defined as follow:
\begin{equation}
    L_{kd} =  \sum_{x \in D} \sum_{i = 1}^{n} -\hat{p}_{i} (x)\log [p_i (x)] +  (1-\hat{p}_{i} (x))\log [ (1-p_i (x))],
\end{equation}
where $\hat{p}_{i} (x)$ and $p_i (x)$ denote the output after a sigmoid activation function of the old model and new model. 

The baseline used in this work, a modified icarl-cnn, is shown in Figure \ref{common2}. 

\section{Proposed approach}
% To alleviate the serious catastrophic forgetting on superclass, we propose a Multi-Teacher Knowledge Distillation (MTKD) strategy, and to further tackle the problem of the uncontrolled prediction number in IIRC, we introduce a simple Top-k prediction restriction mechanism. Notice that both of these two methods can integrate in existing incremental learning method conveniently. 
% We introduce a Multi Teacher Knowledge Distillation loss. It includes a general knowledge distillation loss ${L_{d}}$, and a super-class distillation loss${L_{sd}}$ which is designed to preserve the superclasses knowledge in the first step. Besides, a binary cross-entropy loss ${L_{bce}}$ is used. So, our overall loss function is defined below:
% \begin{equation}
%     L_{overall} = L_{bce} + \lambda L_d + \mu L_{sd}
% \end{equation}
% where ${\lambda}$ and ${\mu}$ are the weights for ${L_{d}}$, ${L_{sd}}$ respectively. 

\subsection{Multi-Teacher Knowledge Distillation}
Existing works~\cite{wu2019large,castro2018end} have shown that learning a new task incrementally would cause catastrophic forgetting in previous similar tasks. In IIRC, this problem is even more serious, which is that superclass knowledge may be replaced by subclass knowledge effortlessly. In this work, we propose a Multi-Teacher Knowledge Distillation strategy to improve superclass discrimination. In MTKD, we ues two teacher models, called as general teacher and superclass teacher to distill knowledge for the student model. Meanwhile, a binary cross-entropy is computed on all the available samples. The overall loss function in the proposed MTKD strategy is as follow:
\begin{equation}
    L = L_{bce} + \lambda L_{gd} + \mu L_{sd},
\end{equation}
where ${\lambda}$ and ${\mu}$ are the balance weights for two knowledge distillation loss ${L_{gd}}$, ${L_{sd}}$ respectively. $L_{bce}$ the binary cross-entropy loss as in Eqn.  (4). $L_{gd}$ is the general teacher distillation loss function and $L_sd$ is the the superclass teacher distillaton loss. $L_{gd}$ and $L_sd$ are described below.

\begin{figure}[ht]
\centering
\includegraphics[scale=0.34]{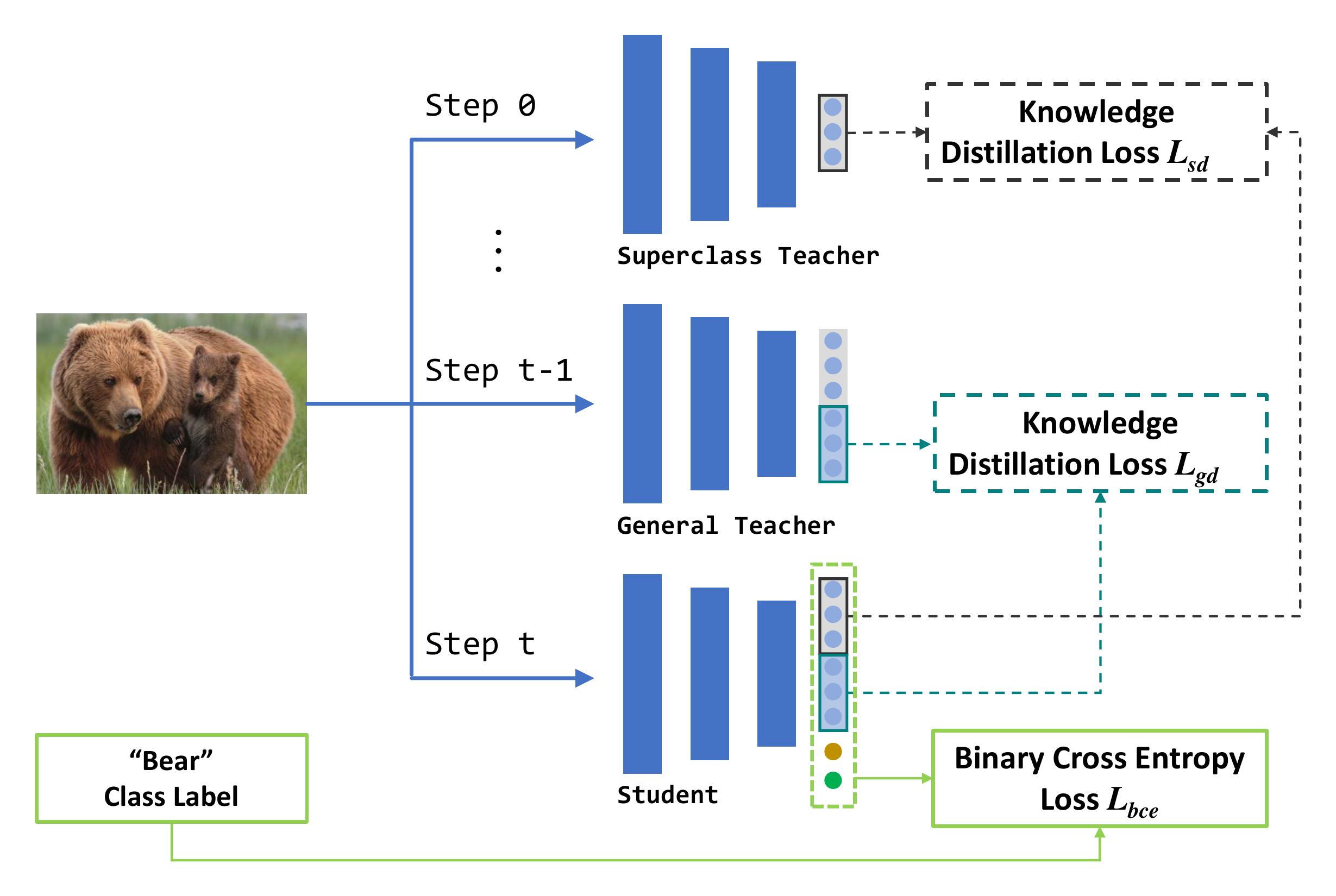}
\caption{In the proposed MTKD strategy, a total of 3 losses are computed for the model update. Superclass distillation loss is conducted on logit of the initial classes between the superclass teacher and the student model. General distillation loss is computed on logits of the classes that appear sequentially, except the initial classes. Binary cross-entropy loss is computed among all the classes.}
\label{MTKD}
\end{figure}

Following conventional incremental learning methods, the model in the last incremental learning step is applied in our method and acts as a general teacher model. The effect of the general teacher is to teach our student model the knowledge of classes appearing subsequently, except for superclasses appearing in the initial incremental step. In our method, the logit distillation loss between the general teacher model and the student model is conducted. Notice that our general teacher knowledge distillation is different from that of conventional incremental learning. In conventional incremental learning methods, existing work affirms the model inherited from the last incremental learning step is proficient in the classification of all the previous classes. We consider that the superclass discrimination may encounter a serious performance drop because superclass knowledge is occupied by their subclass knowledge. Hence, in our MTKD strategy, we do not use the general teacher to distill the knowledge of the superclass appearing in the initial incremental step. Our general teacher distillation loss $L_{gd}$ can be denoted as follow:
\begin{equation}
    L_{gd} =  \sum_{x \in D} \sum_{i = n_0}^{n} -\hat{p}_{i}^g (x)\log [p_i (x)] +  (1-\hat{p}_{i}^g (x))\log [ (1-p_i (x))],
\end{equation}
where $n_0$ denotes the number of classes appearing in the initial incremental step, $\hat{p}_{i}^g (x)$ and $p_i (x)$ denote the outputs after a sigmoid activation function of the general teacher model and student model, respectively.

% In the proposed method, $x_i$ denoted as the training samples from the dataset $ (D_{old}, D_t)$, $D_{old}$ and $D_t$ are denoted as the exemplar dataset of preseverd old samples and traning set of current tasks. $s_0$, $s_{i-1}$ are denoted as total output nodes in task 0, $i-1$ respectively. $q_{i}^y$ and $p_y (x_i)$ are denoted as general teacher's output logits and student's output logits for sample $x_i$ respectively. 

To consolidate the learned superclass knowledge in the initial step, instead of distilling knowledge from the model in the last incremental step, we use the initial model as our superclass teacher model to teach our student model to preserve the superclass knowledge. We consider the superclass knowledge embedded in the initial model are well preserved since superclass knowledge in the initial model does not be occupied by their subclass tasks. In our method, the initial model in the incremental step  (called as superclass teacher) teaches our student model how to distinguish the superclasses appearing in the initial incremental step. Hence, the superclass teacher distillation loss in our MTKD strategy can be denoted as follow:
\begin{equation}
    L_{sd} =  \sum_{x \in D} \sum_{i = 1}^{n_0} -\hat{p}_{i}^s (x)\log [p_i (x)] +  (1-\hat{p}_{i}^s (x))\log [ (1-p_i (x))],
\end{equation}
where $\hat{p}_{i}^s (x)$ and $p_i (x)$ denote the output after a sigmoid activation function of the superclass teacher model and student model. 

% Other definitions are the same as above.

In IIRC, the classification task becomes a multi-label classification task. The conventional softmax-based classification method is not suitable for this task, as the model is required to output more than one prediction. Following IIRC, we add a sigmoid activation function to the last layer of the model. Both the outputs of teacher model $\hat{p}_{i}^g (x)$, $\hat{p}_{i}^s (x)$ and student model $p_i (x)$ are activated by a sigmoid function.

% :
% \begin{equation}
%     p_i (x) =  \frac{1}{1 + exp (-\theta_t^i (x))}
% \end{equation}
% \begin{equation}
%     \hat{p}_{i}^s (x) =  \frac{1}{1 + exp (-\theta_0^i (x))}
% \end{equation}
% \begin{equation}
%     \hat{p}_{i}^g (x) =  \frac{1}{1 + exp (-\theta_{t-1}^i (x))}
% \end{equation}

% where $\theta_t^i (x)$, $\theta_0^i (x)$, $\theta_{t-1}^i (x)$ are denoted as the output for class $i$ of student model, superclass teacher model and general teacher model respectively.

% In addition, our $L_{sd}$ is conducted on the initial superclass output nodes between superclass teacher and the student, while $L_{gd}$ is conducted on all the previous output nodes, except the initial output nodes between general teacher and the student. 

As Figure \ref{MTKD} shows, two models, enact as teachers in our method, the general teacher model uses distillation loss in the previous output nodes except for the initial output nodes. Meanwhile, the superclass teacher model transfers knowledge through the initial class outputs.
 
% In our experiment, we find the model performs poor on superclasses than subclasses. It means the evolution of the weight runs counter to the information of superclasses. It is necessary to consolidate model's memory of superclasses. A simple and effective way to accomplish it is to increase the preserved examples of superclasses. However, this simple idea would increase memory consumption and cannot be implemented when need to consider data privacy issues. In IIRC setup, the first incremental task is always the largest task, and all the classes learned in this step are superclasses. It is in line with human's cognitive process, human always learn some broad concepts at first. In the IIRC setup, the first incremental step imitate this process. Due to human's cognitive process and IIRC setup, We believe that the model trained in the first incremental step is very proficient in the identification of superclasses. Hence, we can use the model trained in the first incremental step, which has abundant superclasses knowledge to supplement the lack of superclass knowledge in the subsequent model. Knowledge distillation is an very successful technique for knowledge transfer between two models. As the figure 3[] shows, two models act as teachers in our method, the general teacher model uses distillation loss in the previous output nodes except the initial output nodes. Meanwhile, the superclass teacher model transfer knowledge through the initial output nodes.

In our method, the model in the initial step is preserved and acts as a teacher to distill its knowledge to the subsequent model. Hence, our method would increase storage consumption. A naive strategy to alleviate catastrophic forgetting in incremental learning is consuming much storage space to preserve old exemplars for training.  However, in our experiments, we compare the performance between our method and the case of using the same storage consumption to store old samples for training. The experimental results show, under the same circumstances, our method performs far better than the latter case. 

% Where ${x_i}$ is our training samples in the ${t_{th}}$ incremental step, ${x_i \in D_t}$. ${D_t}$ includes samples from new tasks data and rehearsal samples of previous tasks. $s$ represents the total superclasses in the initial step (i.e 10 for IIRC-CIFAR, 63 for IIRC-ImageNet). ${q_i^y}$ is the superclass teacher's prediction for category $y$ while the input is $x_i$. ${p_y (x_i)}$ is the prediction of the current student model. Note that both the prediction of teacher and student is after a sigmoid activation function. For our model $M$, we can split the parameters of model $M$ into a feature extraction part and a variable classifier weight vectors. We denote the parameters of feature extraction part as $\phi$, and the weight vectors as $w_1, ..., w_t \in R^d$. For the teacher models, the weight vectors and feature extraction part are denoted as $w_1^s, ..., w_t^s \in R^d$, $w_1^g, ..., w_t^g \in R^d$ for $\phi^s$, $\phi^g$ respectively. So the ${q_i^y}$ and ${p_y (x_i)}$ can be defined as bellow:

Our Multi-Teacher Knowledge Distillation is designed as a universal strategy for all the existing incremental learning methods. Hence, the general teacher in our method can be any form. For instance, in icarl-MTKD, the general teacher acts as the old model, distilling logits between old models and new models. In PODNet-MTKD, the general teacher acts as the old model, distilling both logits and attention feature map between models. Our experiments show that our MTKD can combine with existing methods and show great performance improvement.

\subsection{Top-k prediction restriction}
The introduction of the superclass teacher in MTKD strategy could cause a redundant prediction problem, which means the final activated prediction number exceeds the maximum number of Labels per image. In IIRC, the final predictions with score greater than 0.5 would be activated. The prediction mechanism in IIRC can be defined as:
\begin{equation}
    \begin{cases}
        o_i (x) =True,  p_i (x)>0.5 \\
        o_i (x) =False, p_i (x)<=0.5,
    \end{cases}
\end{equation}
where $o_i (x)$ represents the final prediction of sample $x$,  $o_i (x) =True$ means the the model considers sample $x$ belongs to class $i$.

To avoid performance drop, we propose a simple Top-k prediction restriction to reduce redundant predictions. In the Top-k prediction restriction mechanism, for the final prediction scores $p (x)$, $ (p_1, p_2, ..., p_{n+m})$, the largest k elements are retained and make a new Top-k prediction score $\bar{p} (x)$. We adopt final predictions with where Top-k prediction score $\bar{p} (x)$ greater than 0.5 as the final predictions. It can denoted as follow:
\begin{equation}
    \begin{cases}
        o_i (x) =True,  \bar{p}_i (x)>0.5 \\
        o_i (x) =False, \bar{p}_i (x)<=0.5.
    \end{cases}
\end{equation}

\section{Experiments}

\subsection{Evaluation Metrics}
Conventional class incremental learning uses average incremental accuracy~\cite{rebuffi2017icarl} of all the classes as their metric, but it is not suitable for IIRC. The Jaccard similarity  (JS) is a popular metric in the multi-label classification task, which is defined as:
\begin{equation}
JS = \frac{1}{n_k} \sum_{i=1}^{n_k} {\frac{\left |Y_{ki}\cap \hat{Y}_{ki} \right |}{\left |Y_{ki}\cup\hat{Y}_{ki}\right |}}.
\end{equation}

In IIRC, the ratio of true positive over the sum of the true positives and false positives is used to weight the JS to further penalize the uncertainty model. Hence in IIRC, they propose a precision-weighted Jaccard Similarity  (pw-JS)~\cite{abdelsalam2020iirc} as follows:
\begin{equation}
  R_{jk} = \frac{1}{n_k} \sum_{i=1}^{n_k} {\frac{\left |Y_{ki}\cap \hat{Y}_{ki} \right |}{\left |Y_{ki}\cup\hat{Y}_{ki} \right |}} \times {\frac{\left |Y_{ki}\cap \hat{Y}_{ki}\right |}{\left |\hat{Y}_{ki}\right |}},
\end{equation}
where $j$ denotes current increment step, $k$ denotes the previous increment step $ j\geq$k. ${Y}_{ki}$, $\hat{Y}_{ki}$ are true label and the model predictions for the $i_{th}$ sample in the $k_{th}$ increment step. 

To get the performance of all the previous tasks, the final evaluation metric is the average precision-weighted Jaccard similarity over all the classes the model has seen. Notice that in each task, although the model could see only one label for all the classes, the model is required to predict all the correct labels in the test time. The final metric on a total of n test samples is below:
\begin{equation}
 R_{j} = \frac{1}{n} \sum_{i=1}^{n} {\frac{\left |Y_{i}\cap \hat{Y}_{i}\right |}{\left |Y_{i}\cup\hat{Y}_{i}\right |}} \times {\frac{\left |Y_{i}\cap \hat{Y}_{i}\right |}{\left |\hat{Y}_{i}\right |}}.
\end{equation}

\begin{figure*}[htbp]
\centering
\subfigure{
\begin{minipage}[t]{0.3\linewidth}
\centering
\includegraphics[width=1.8in]{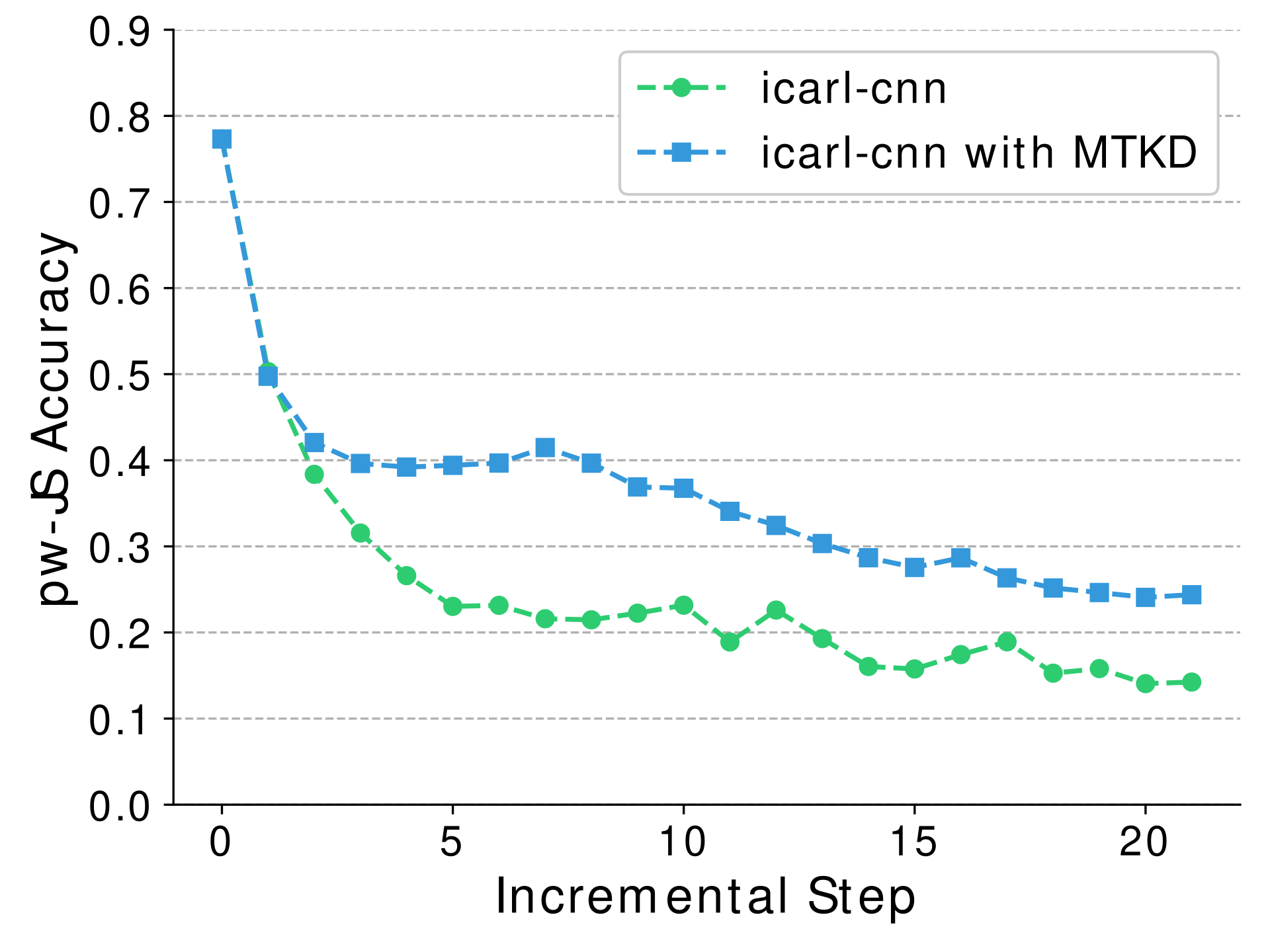}
%\caption{fig1}
\end{minipage}%

\begin{minipage}[t]{0.3\linewidth}
\centering
\includegraphics[width=1.8in]{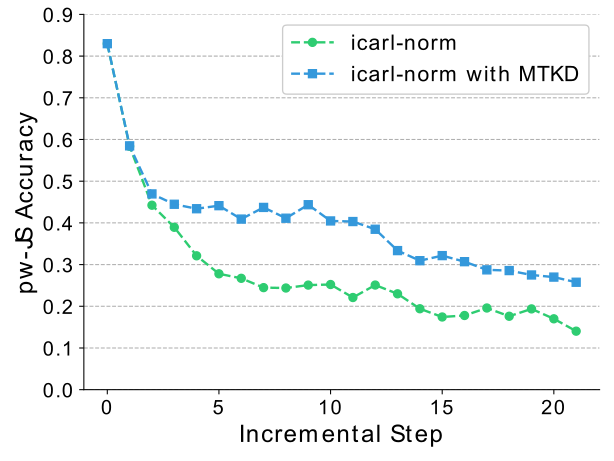}
%\caption{fig1}
\end{minipage}%

\begin{minipage}[t]{0.3\linewidth}
\centering
\includegraphics[width=1.8in]{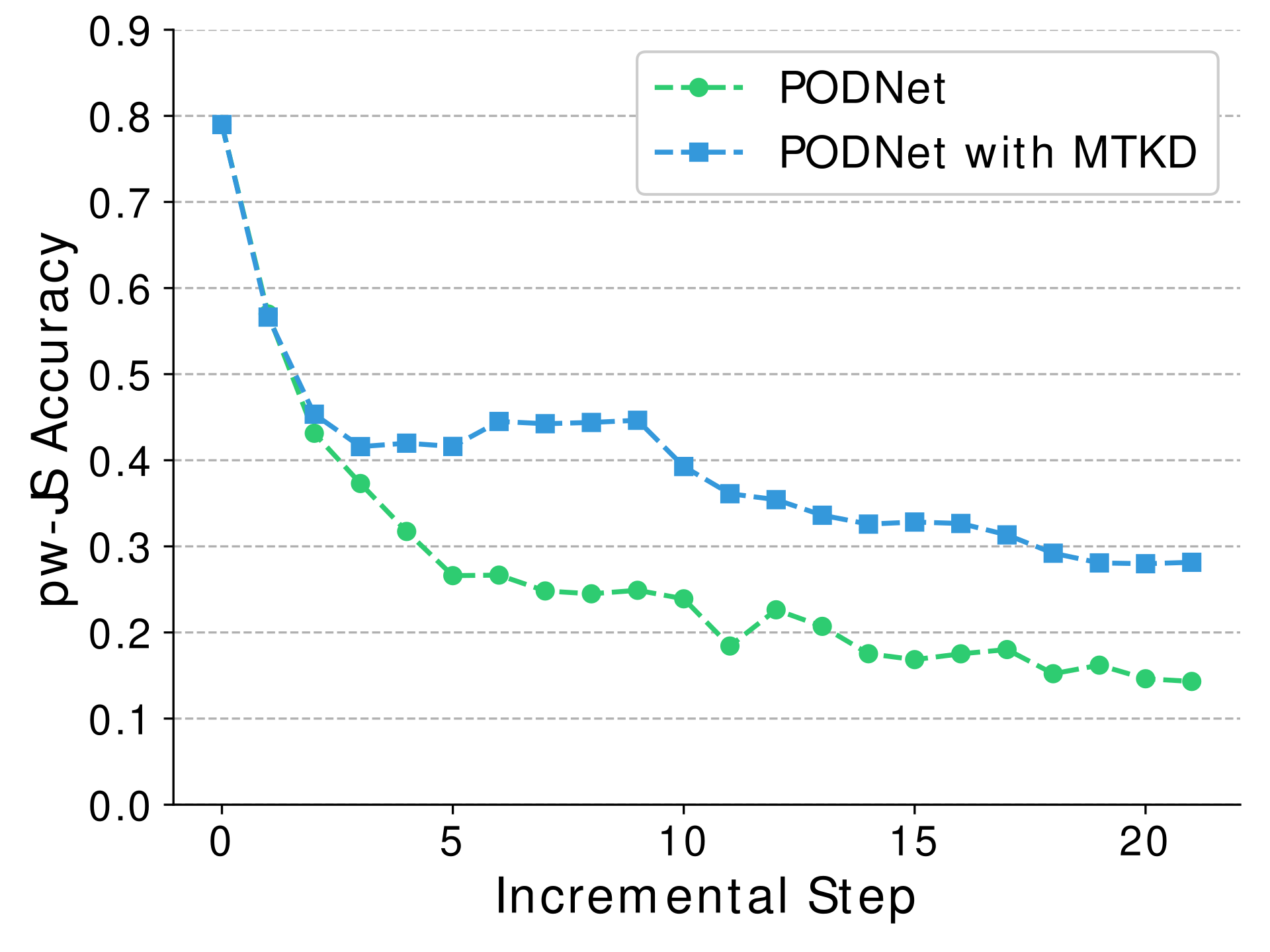}
%\caption{fig1}
\end{minipage}%
}%

% \subfigure{
% \begin{minipage}[t]{0.3\linewidth}
% \centering
% \includegraphics[width=2.1in]{ICARL3.png}
% %\caption{fig1}
% \end{minipage}%

% \begin{minipage}[t]{0.3\linewidth}
% \centering
% \includegraphics[width=2.1in]{ICARL_norm3.png}
% %\caption{fig1}
% \end{minipage}%

% \begin{minipage}[t]{0.3\linewidth}
% \centering
% \includegraphics[width=2.1in]{podnet_3.png}
% %\caption{fig1}
% \end{minipage}%
% }%

\subfigure{
\begin{minipage}[t]{0.3\linewidth}
\centering
\includegraphics[width=1.8in]{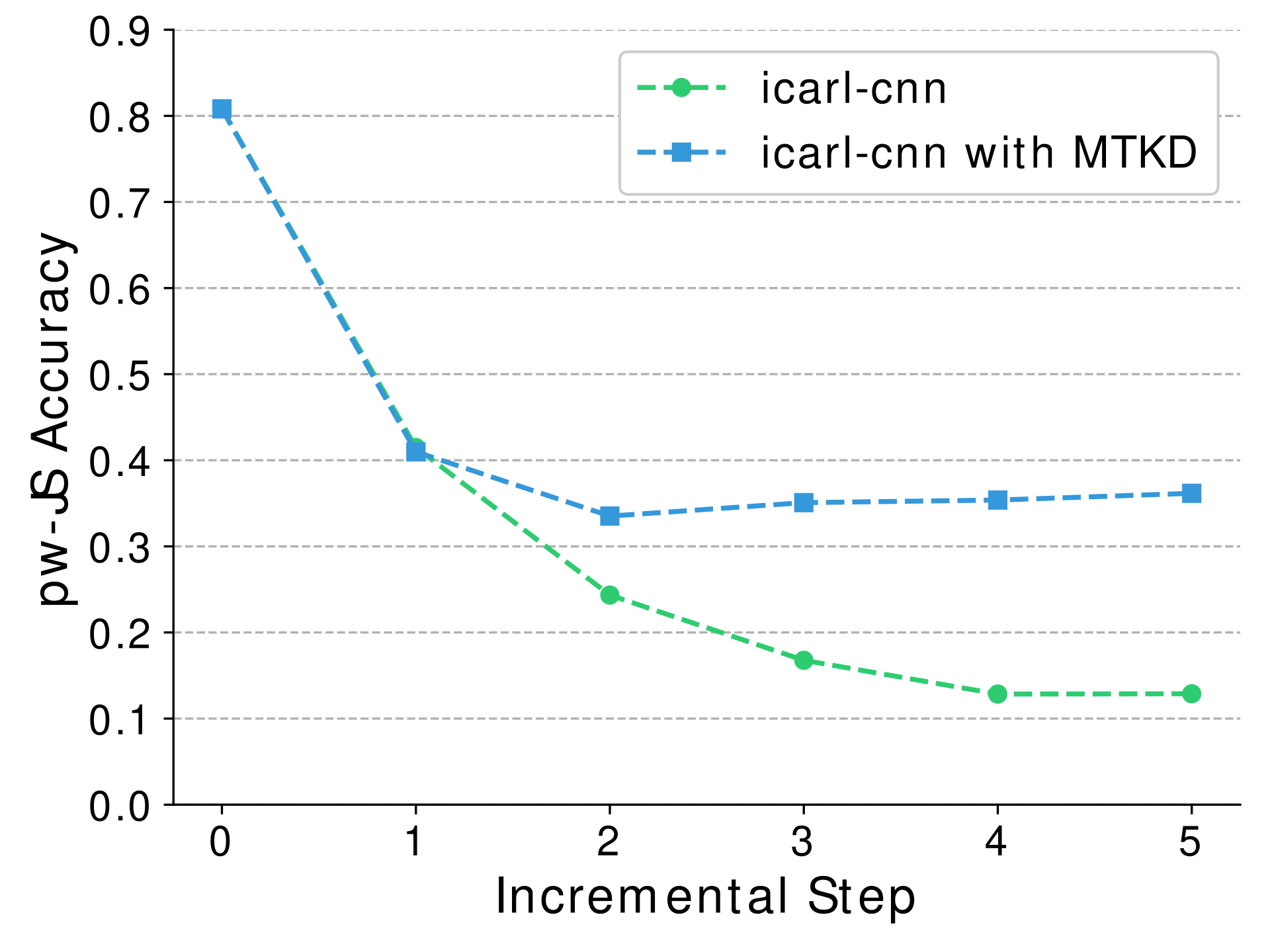}
%\caption{fig1}
\end{minipage}%

\begin{minipage}[t]{0.3\linewidth}
\centering
\includegraphics[width=1.8in]{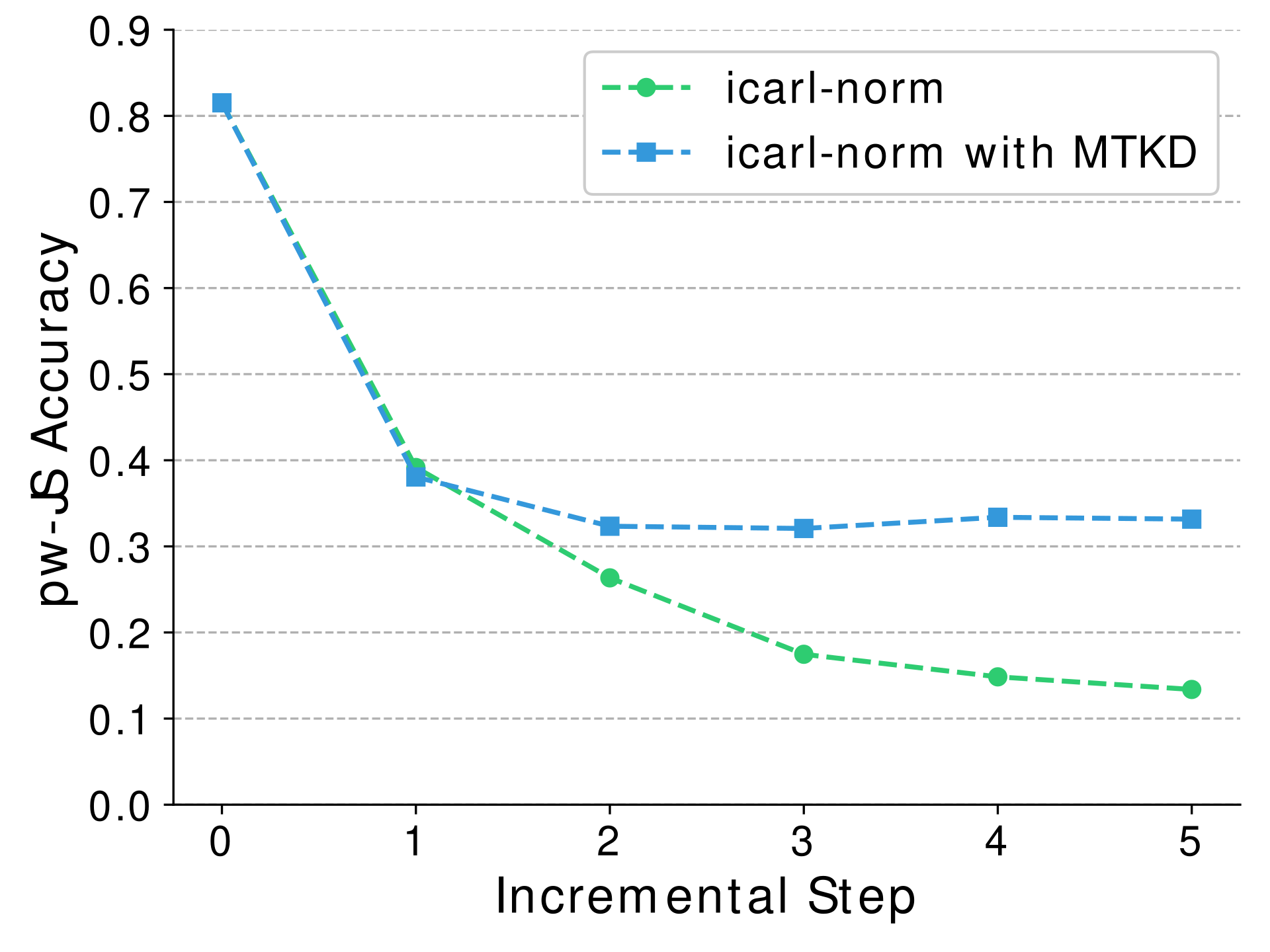}
%\caption{fig1}
\end{minipage}%

\begin{minipage}[t]{0.3\linewidth}
\centering
\includegraphics[width=1.8in]{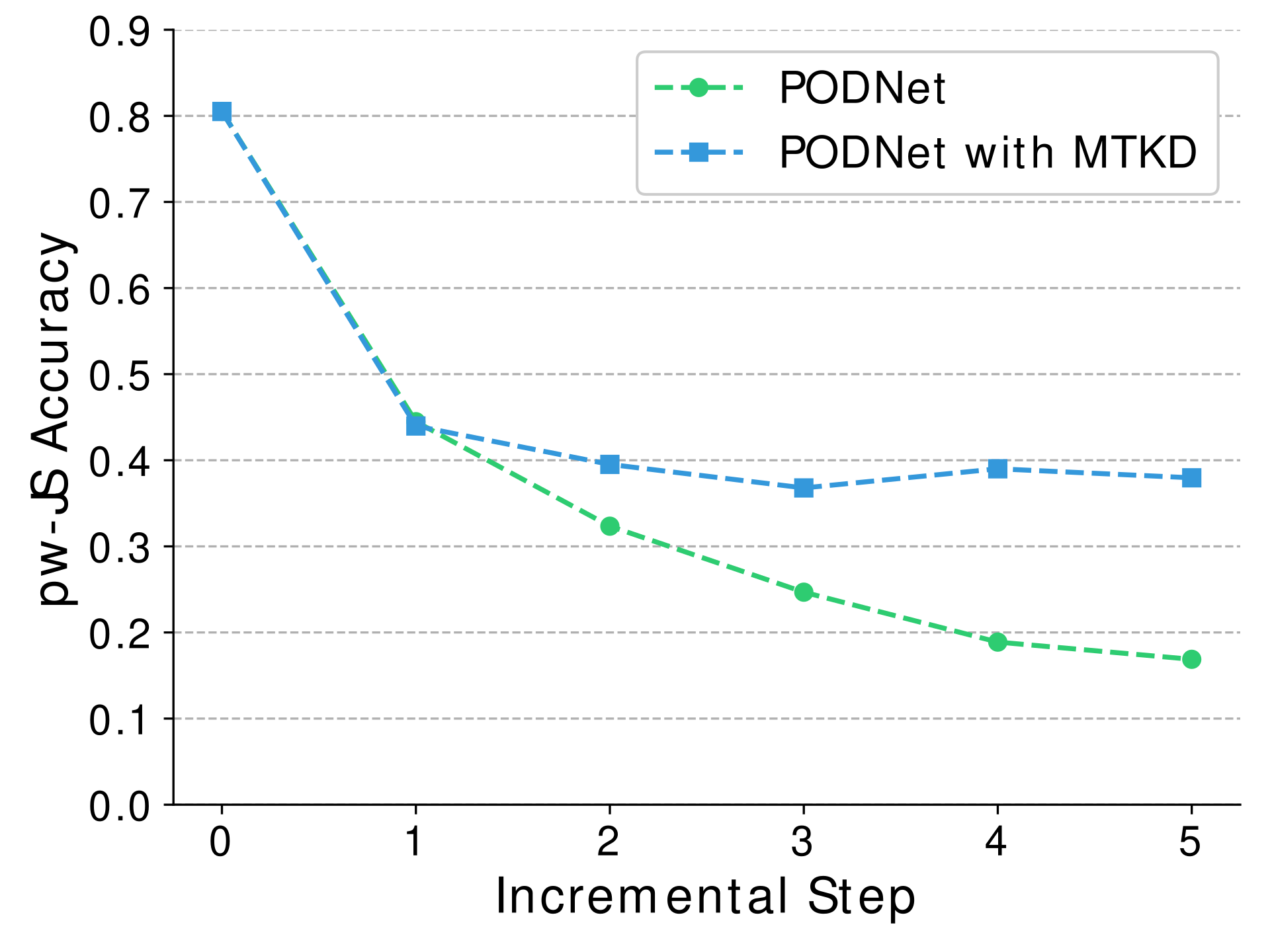}
%\caption{fig1}
\end{minipage}%
}%
\centering
\caption{Comparisons of pw-JS accuracy between our MTKD strategy and state-of-the-art. Top and down rows show 3 baselines and their combination with our MTKD strategy on IIRC-CIFAR100 under 22 incremental steps and on the IIRC-ImageNet120 under 6 steps.}
\label{performance_mtkd}
\end{figure*}

\begin{table}
    \caption{\textbf{Comparisons of average pw-JS accuracy on IIRC-CIFAR100 under 5 classes  (22 steps), 10 classes  (11 steps) and on IIRC-ImageNet120 under 20 classes  (6 steps) with our proposed MTKD strategy and state-of-the-art.}}
    \begin{minipage}{0.98\linewidth}
        \centering
        
        \label{table:snetv2}
        \resizebox{1\textwidth}{!}{
         \begin{tabular}{l|ll|l}
          \toprule

          \multirow{2}*{Methods} &  \multicolumn{2}{c|}{CIFAR100} & ImageNet120 \\ 
          \cline{2-4} 
        %   \cline{4}
          ~ & \makecell[c]{5 classes}  &\makecell[c]{10 classes} & \makecell[l]{20 classes}\\
          \midrule 
          icarl-cnn    & 14.2 & 17.6 & 12.9\\ 
            ~~~~ + MTKD  & \textbf{24.3} \textcolor{red}{\footnotesize{+10.1}} & \textbf{25.1} \textcolor{red}{\footnotesize{+7.5}}& \textbf{36.1} \textcolor{red}{\footnotesize{+23.2}} \\ 
            % ~~~~ + k-MTKD  & \textbf{25.6} \textcolor{red}{\footnotesize{+11.4}} & \textbf{25.9} \textcolor{red}{\footnotesize{+8.3}}& \textbf{38.6} \textcolor{red}{\footnotesize{+25.7}} \\ 
          \midrule
          \midrule 
          icarl-norm        & 14.1 & 21.1 & 13.4 \\ 
            ~~~~ + MTKD  & \textbf{25.8} \textcolor{red}{\footnotesize{+11.7}} & \textbf{27.1} \textcolor{red}{\footnotesize{+6.0}} & \textbf{33.1} \textcolor{red}{\footnotesize{+19.7}} \\ 
          \midrule
          \midrule
        %   LUCIR             & 0 & 0  \\ 
        %   LUCIR   +  \textbf{MTKD}     & 0 & 0  \\ \hline 
          PODNet             & 14.3 & 16.2 & 16.8  \\ 
          ~~~~   + MTKD    & \textbf{28.1} \textcolor{red}{\footnotesize{+13.8}} & \textbf{27.4} \textcolor{red}{\footnotesize{+11.2}}& \textbf{37.9} \textcolor{red}{\footnotesize{+21.1}} \\
          \bottomrule
        \end{tabular}
        }
    \end{minipage}
\label{performance_table}
\end{table}

\subsection{Implementation details}
We evaluate the methods on IIRC-ImageNet120 and IIRC-CIFAR100. There are a total of 115 classes in IIRC-CIFAR100, among them, 15 classes are superclasses, 77 classes are subclasses with a parent class, the other 23 classes are subclasses with no parent. In the IIRC-ImageNet120, a total of 120 classes are introduced, this number contains 20 superclasses and 100 subclasses under a parent. Following IIRC, we use reduced ResNet-32~\cite{he2016deep} for IIRC-CIFAR100, and ResNet-50~\cite{he2016deep} for IIRC-ImageNet120 and we use SGD as our optimizer, as it is more suitable for incremental learning~\cite{mirzadeh2020understanding}. The initial learning rate is set as 1.0, 0.5 for IIRC-CIFAR100 and IIRC-ImageNet120, respectively, and reduces to 1/10 on plateau. We train the network on a total of 100 epoch and 80 epoch per task for IIRC-CIFAR100 and IIRC-ImageNet120. The loss weight in our MTKD strategy, ${\lambda}$ and ${\mu}$ are both set as 0.5, and the max number of predictions k is 2. There is a fix memory consumption of 20 samples per class. We use herding~\cite{welling2009herding} strategy and random selection to select exemplars for IIRC-CIFAR100 and IIRC-ImageNet120, respectively.

\subsection{Baselines}
In our experiments, we select some representative method for class incremental learning, icarl-cnn~\cite{rebuffi2017icarl}, icarl-norm~\cite{abdelsalam2020iirc}, PODNet~\cite{douillard2020podnet} as our baseline. To illustrate the effectiveness of our MTKD strategy, we integrate these methods into our MTKD strategy and observe the performance improvement. The final evaluation criterion is the pw-JS metric proposed in IIRC.

\begin{figure*}[htbp]
\centering
	\begin{tabular}{cccc}
	\centering
	\includegraphics[width=1.5in]{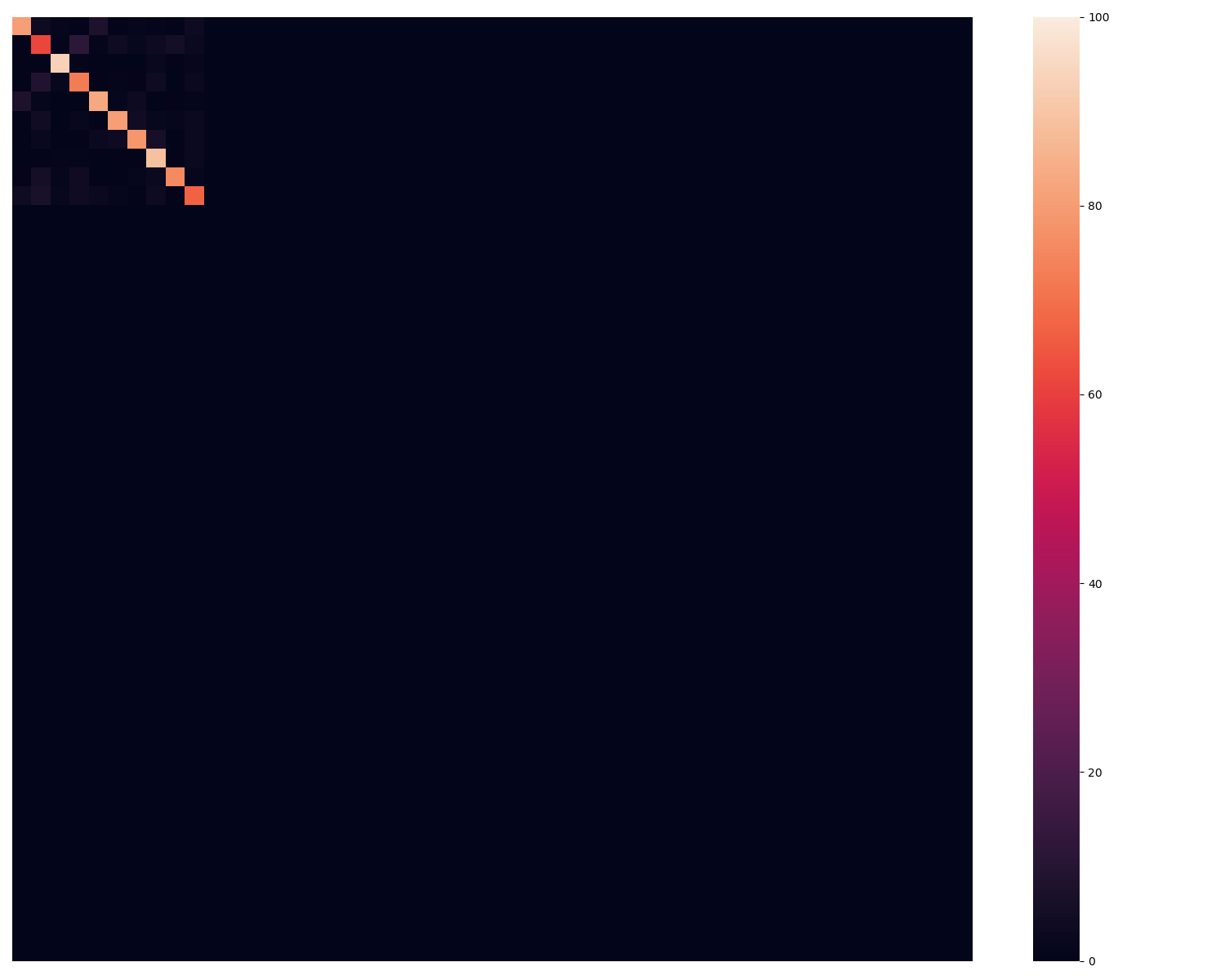}&
	\includegraphics[width=1.5in]{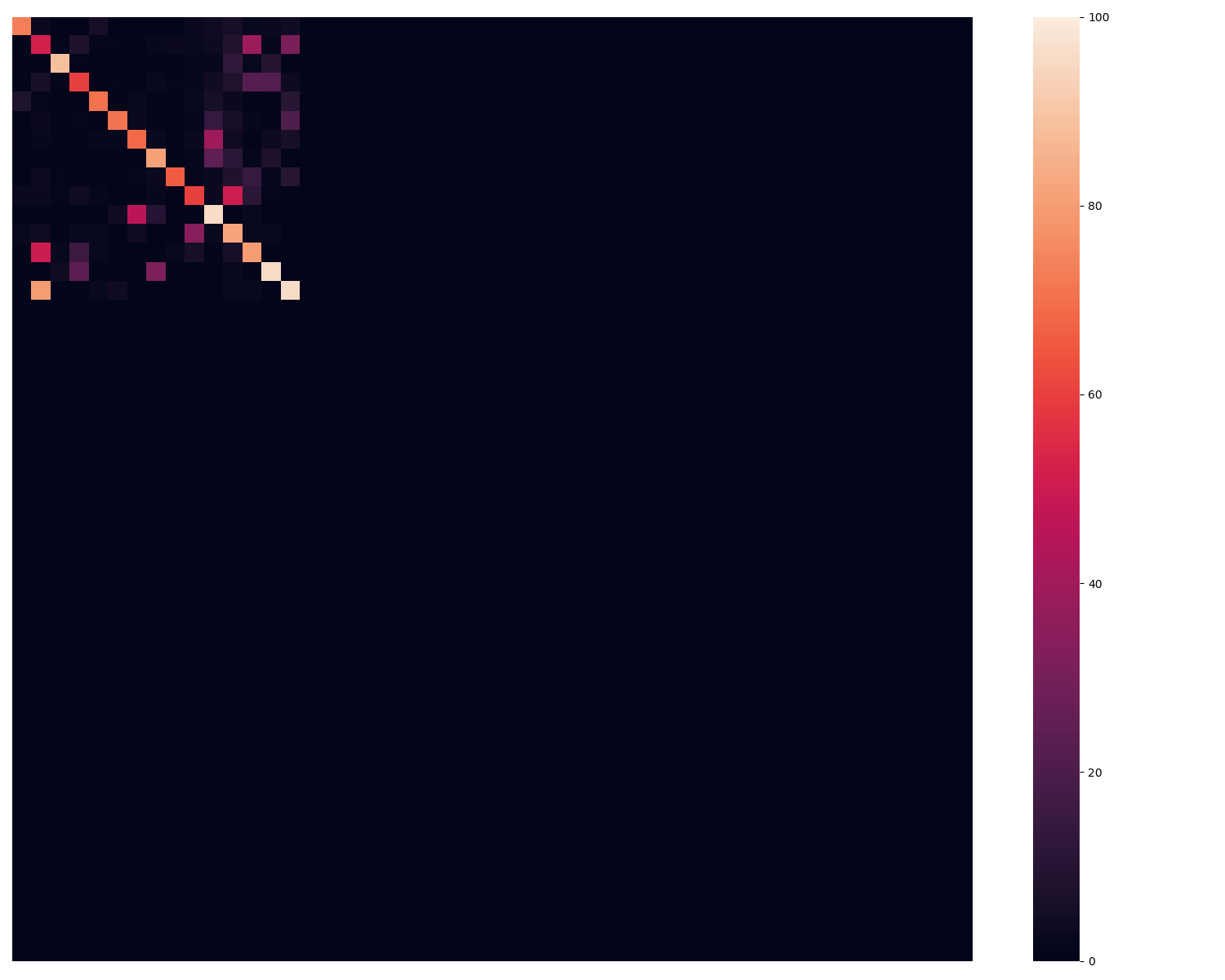}&
	\includegraphics[width=1.5in]{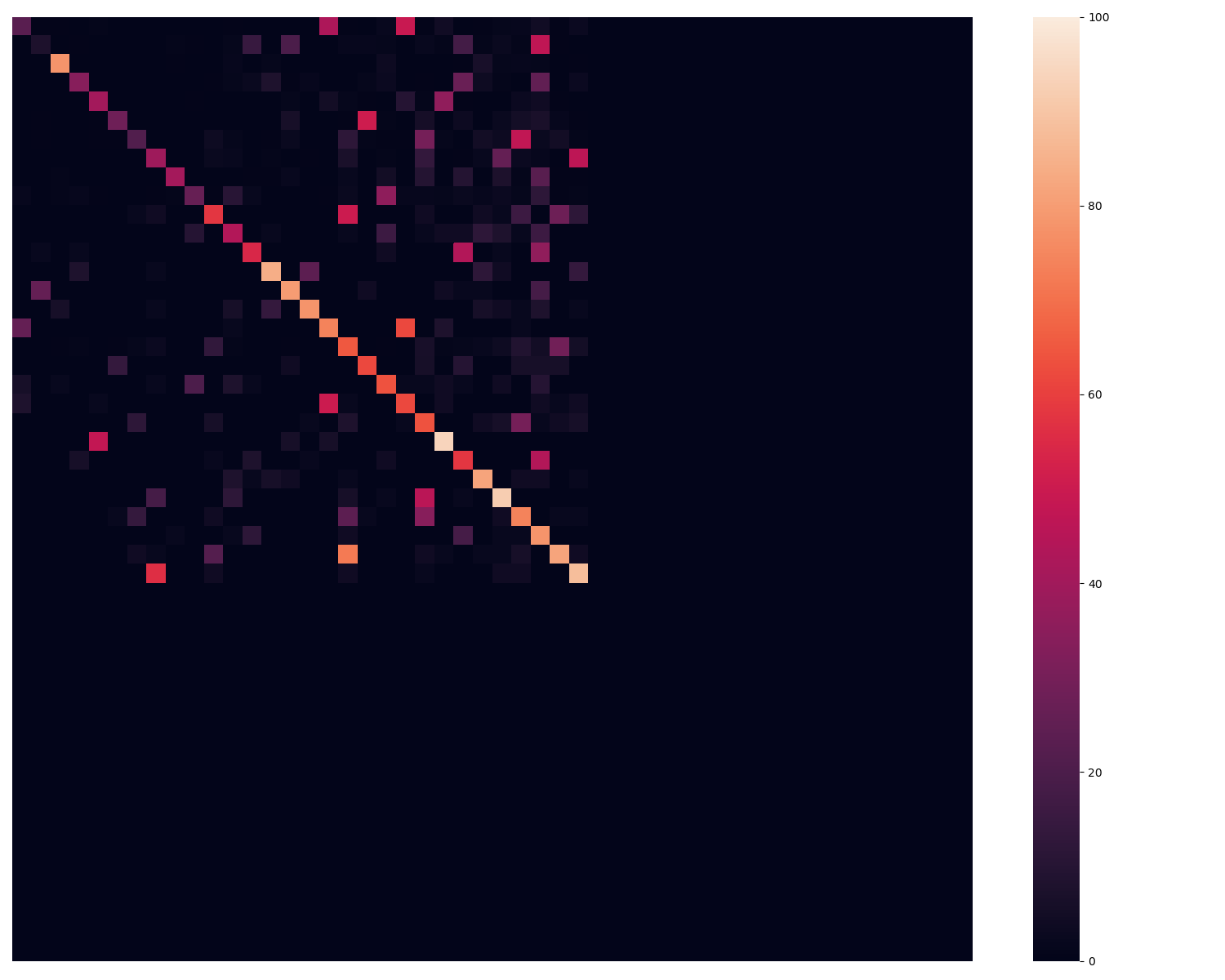}&
	\includegraphics[width=1.5in]{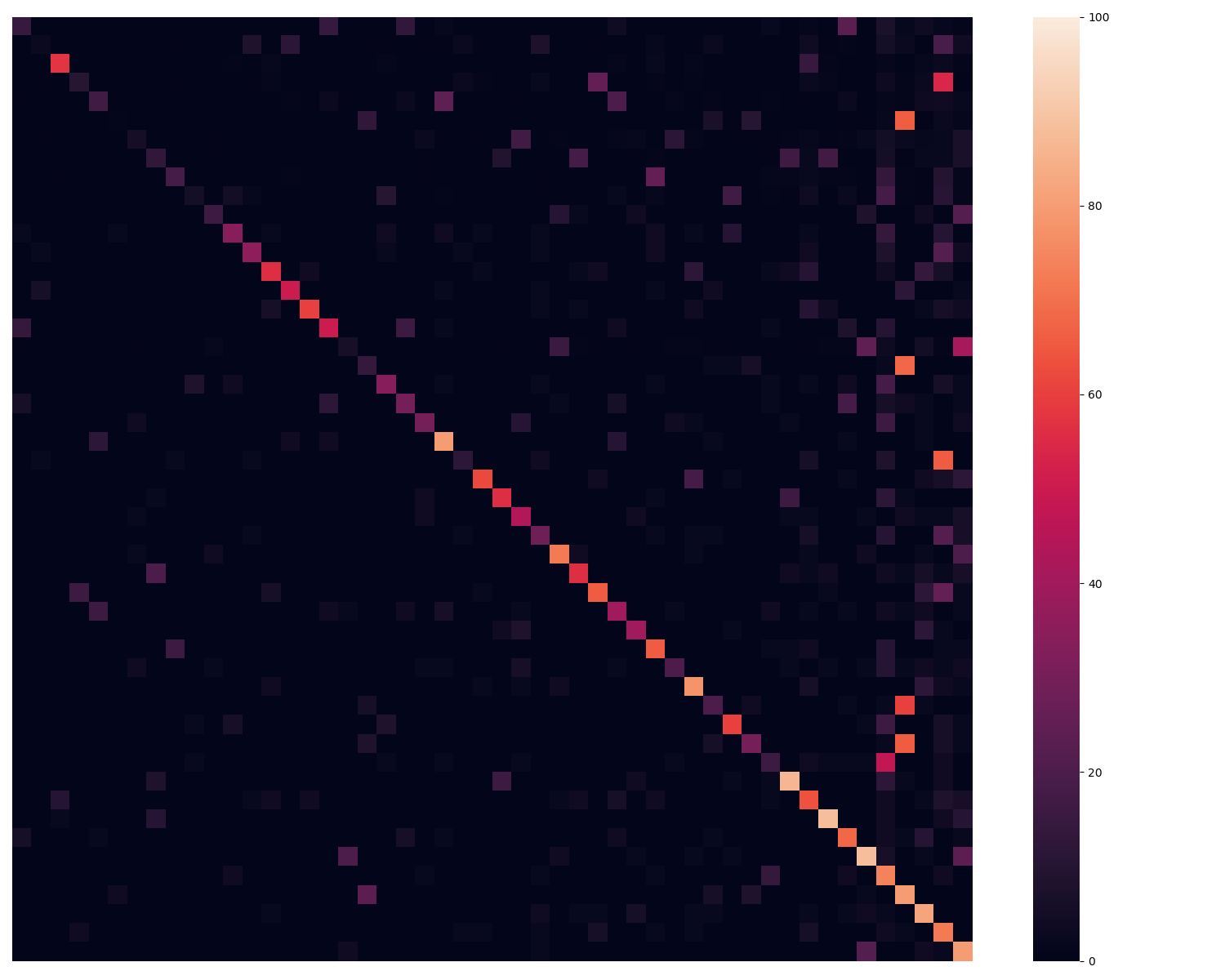}\\
	\scriptsize{(a) After task0} & \scriptsize{(b) After task1} & \scriptsize{(c) After task4} & \scriptsize{(d) After task8} \\
\end{tabular}		
\begin{tabular}{cccc}
	\includegraphics[width=1.5in]{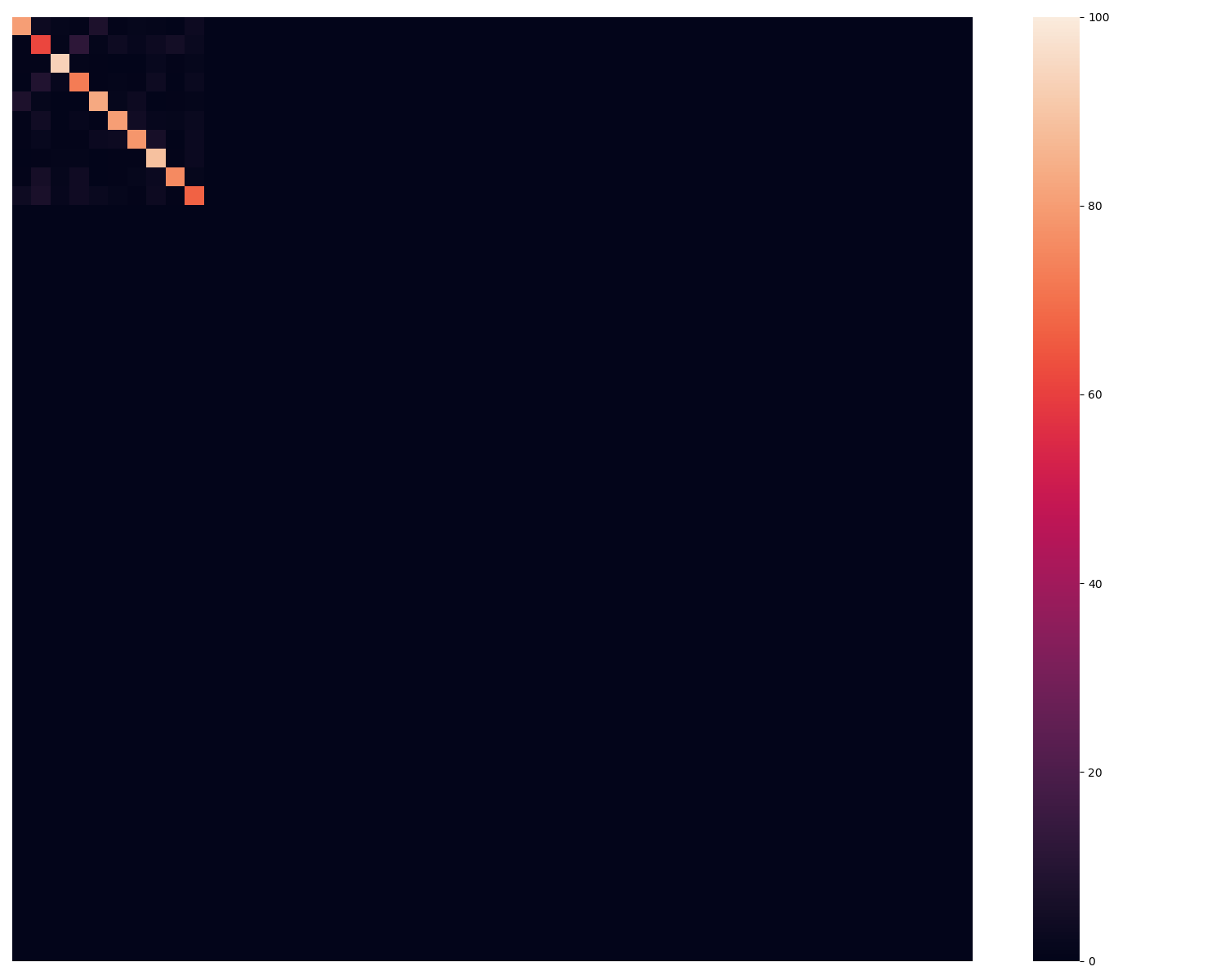}&
	\includegraphics[width=1.5in]{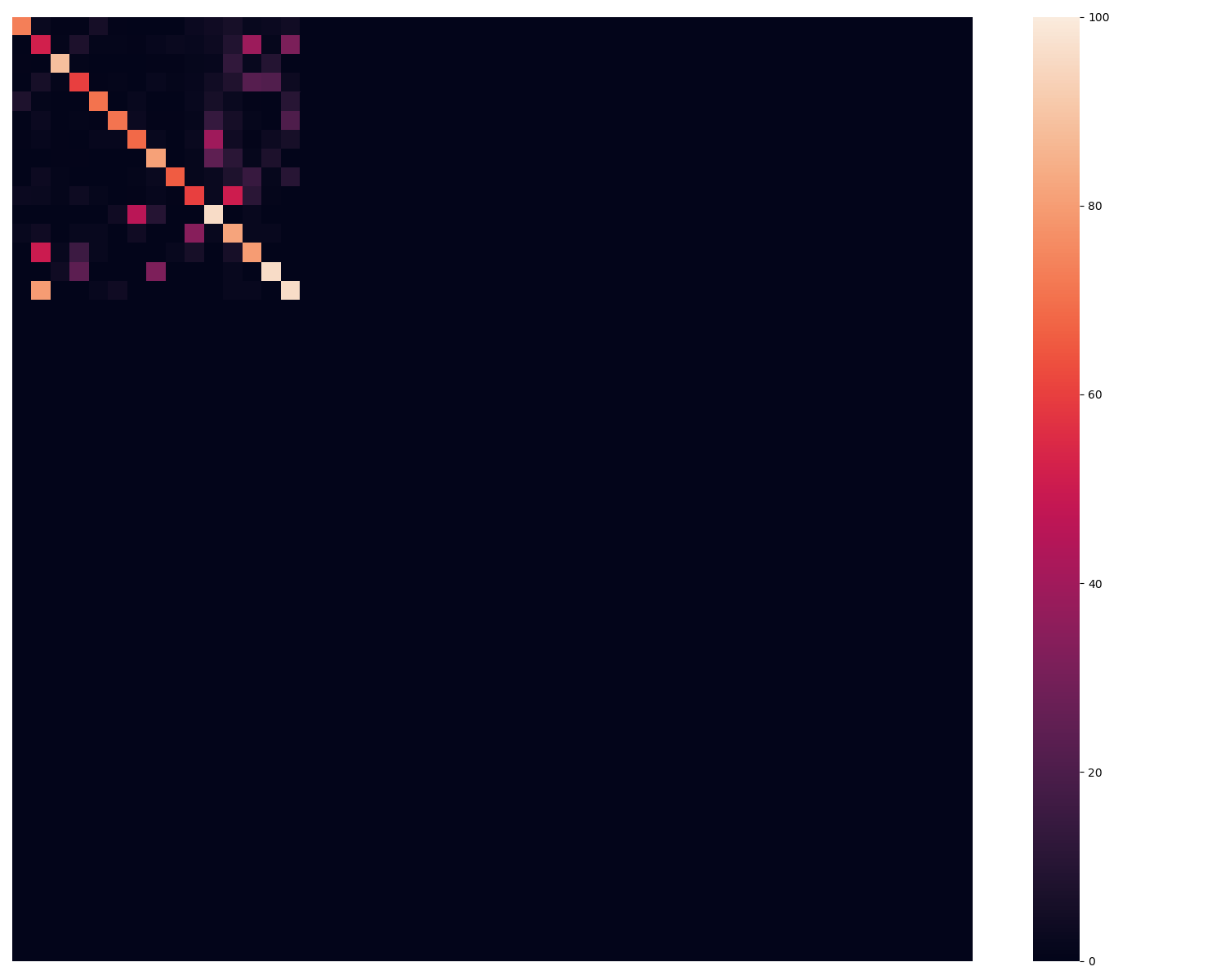}&
	\includegraphics[width=1.5in]{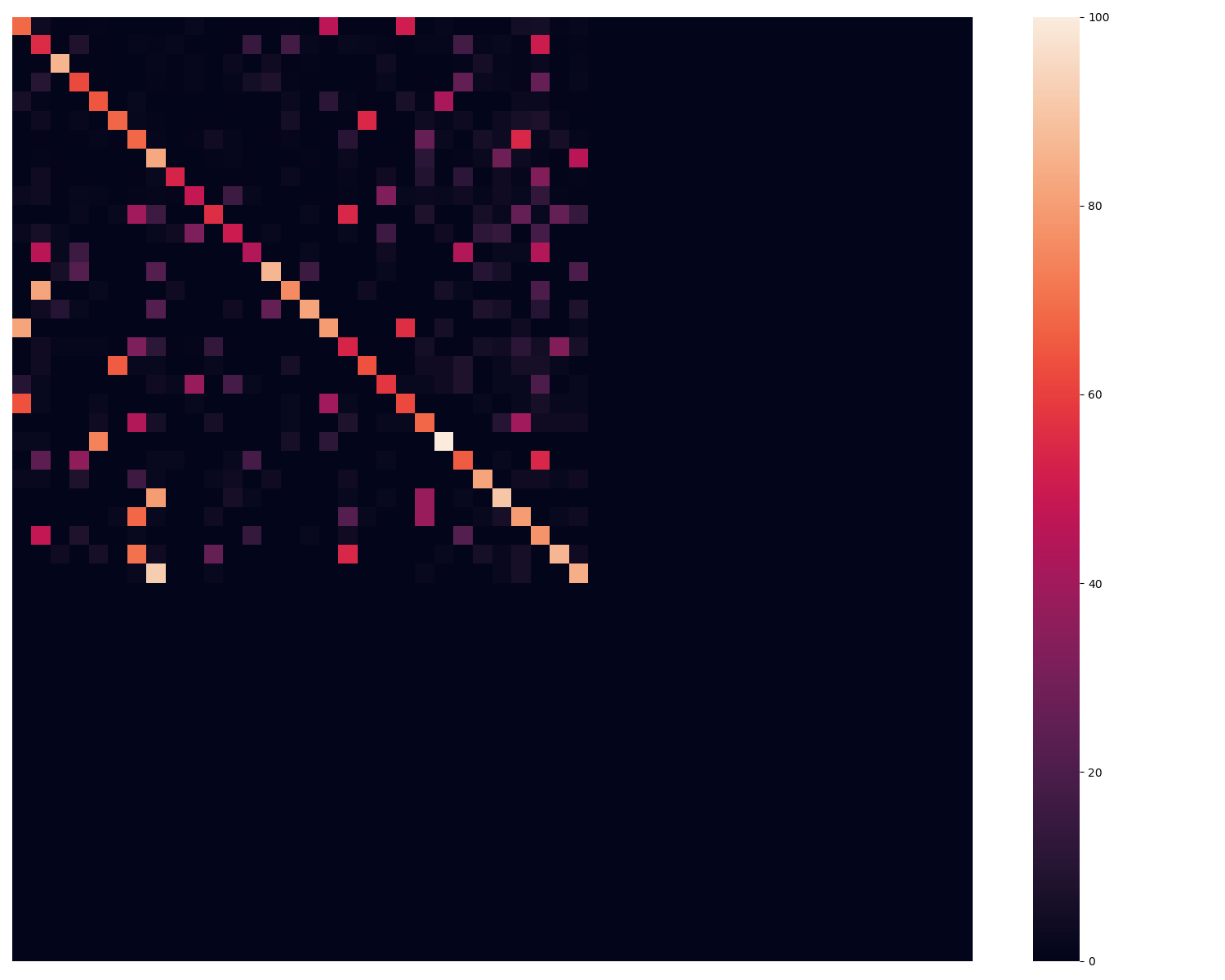}&
	\includegraphics[width=1.5in]{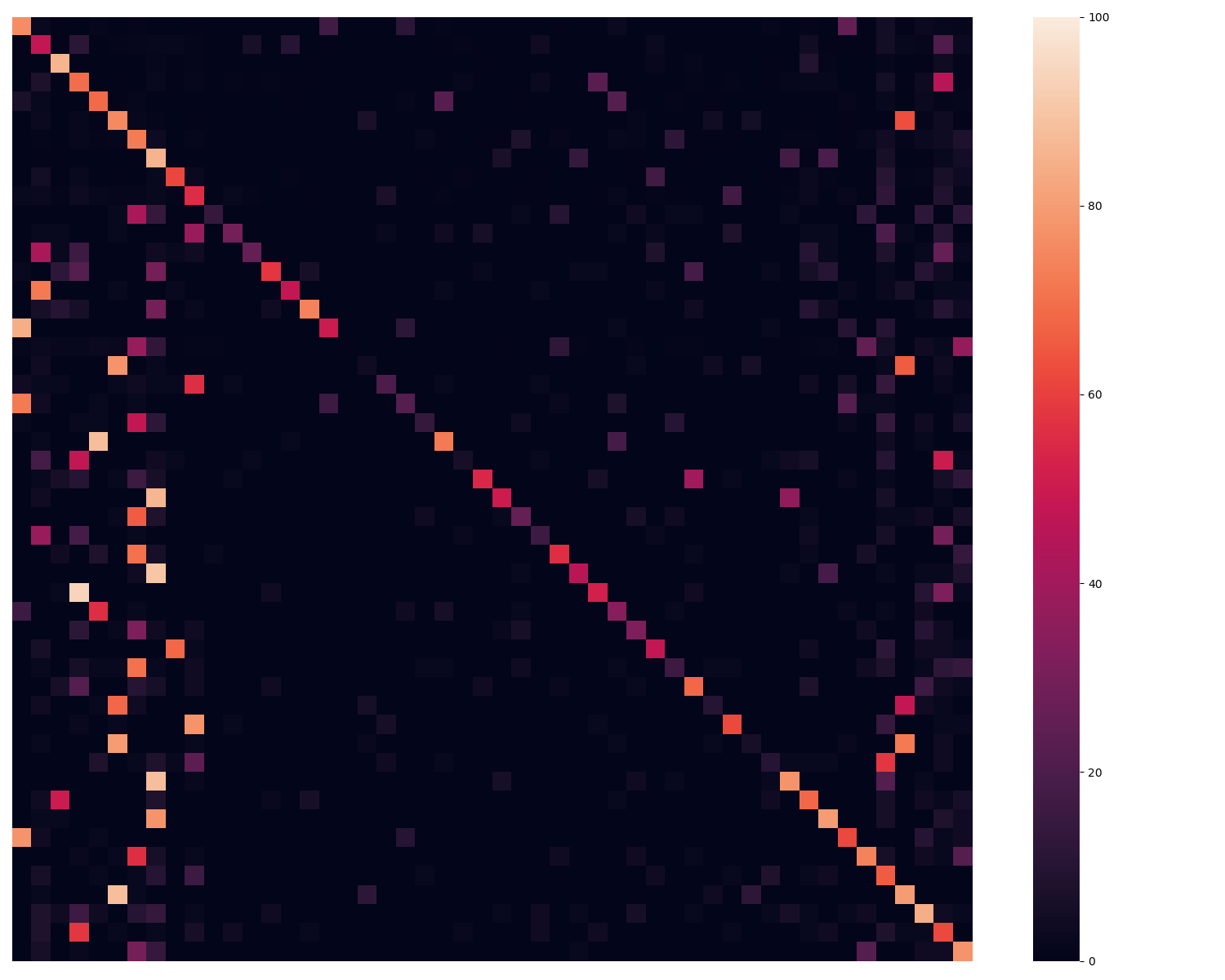}\\
	\scriptsize{(e) After task0} & \scriptsize{(f) After task1} & \scriptsize{(g) After task4} & \scriptsize{(h) After task8} \\
\end{tabular}		
\centering
\caption{Confusion matrix of icarl-cnn (a)(b)(c)(d) and icarl-cnn with MTKD (e)(f)(g)(h) after introducing task0, 1, 4, 8 of IIRC-CIFAR100 respectively. The y-axis is the correct label, and the x-axis is the model prediction.}
\label{confusion_mtkd}
\end{figure*}

\subsection{Evaluation on IIRC-CIFAR100}
IIRC-CIFAR100 has 15 superclasses, 77 subclasses with a parent class, and the other 23 subclasses with no parent. In incremental learning, the different number of incremental steps has a great impact on the accuracy of incremental learning methods. To verify the robustness of our proposed method, we have done experiments on two different configurations for IIRC-CIFAR100. One configuration starts at 10 superclasses and increases 5 classes each step, while the other starts at 15 classes, including 10 superclasses and 5 subclasses, and increases 10 classes each step. Table \ref{performance_table} shows our proposed MTKD strategy can combine with existing methods conveniently and improve the performance by a large margin under different incremental learning steps. In our experiments, the maximum gain is in PODNet with MTKD, which can get 13.8\% performance improvement. Besides, we find the proposed MTKD strategy works well in a long incremental step configuration. Our MTKD strategy has an average performance improvement of 11.8 \% in a 22 incremental step setting  (5 classes), while the average improvement is 8.2 \% in  the 11 incremental step setting  (10 classes). This is due to that the superclass teacher in our MTKD strategy is the initial model in the incremental step and can help our student consolidate the knowledge of initial classes, which would suffer serious catastrophic forgetting in a long incremental step setting. Figure \ref{performance_mtkd} shows the pw-JS accuracy curve across incremental steps under the configuration of 22 incremental steps. All of the methods with our MTKD strategy can exceed the original method after the $2_{th}$ step. This is because in the $1_{th}$ step, our MTKD strategy is the same as the original method, as the superclass teacher and general teacher in our MTKD strategy are identical.

\begin{figure}[ht]
\centering
\includegraphics[scale=0.45]{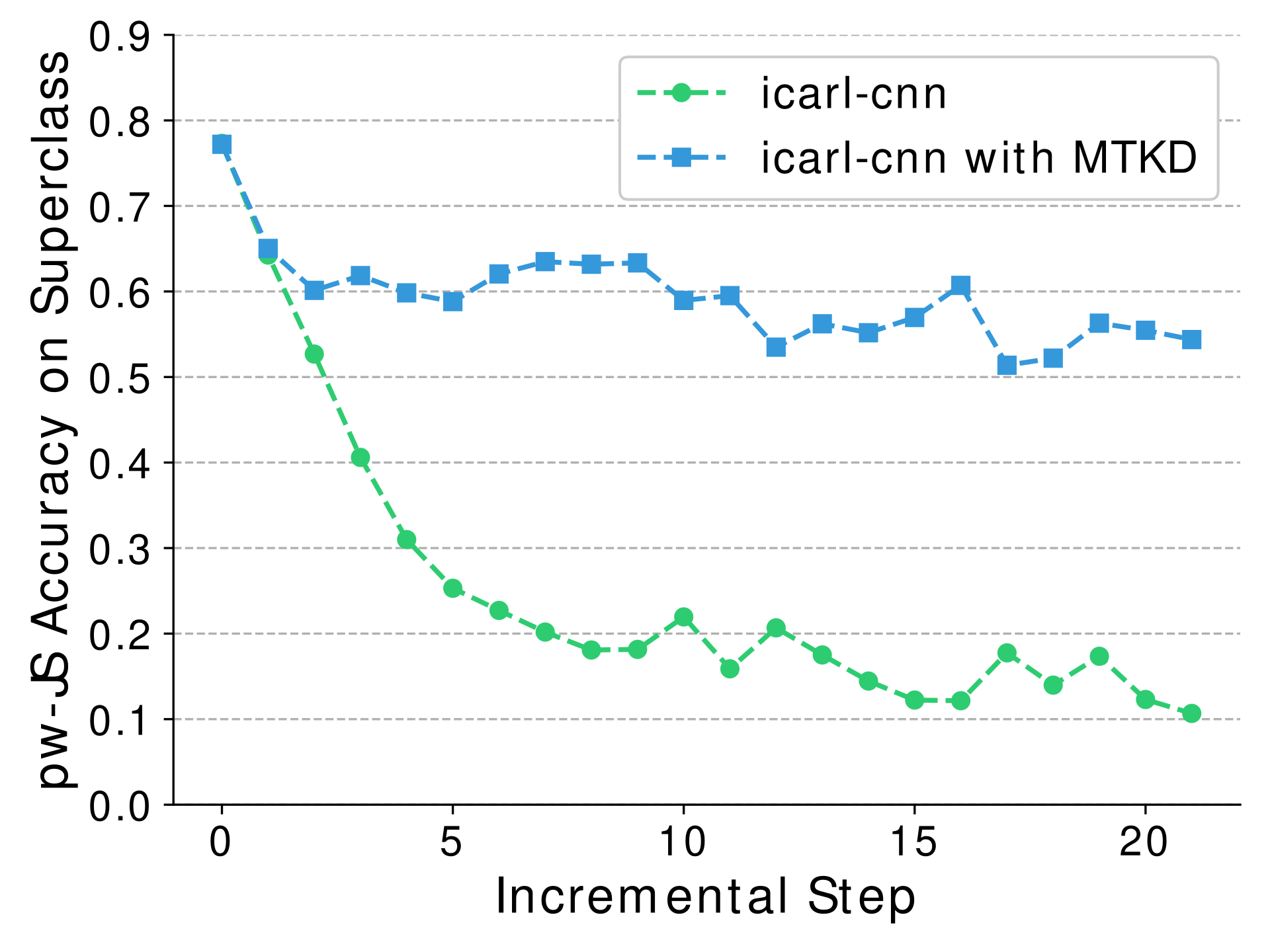}
\caption{Comparison of pw-JS on superclass between icarl-cnn and icarl-cnn with MTKD. Our MTKD strategy alleviates catastrophic forgetting by significantly improve the discrimination of the superclass.}
\label{superclass}

\end{figure}

\subsection{Evaluation on IIRC-ImageNet120}
In this section, we compare our MTKD strategy with other state-of-the-art methods on IIRC-ImageNet120, which is proposed in this work, IIRC-ImageNet120 has a total of 6 incremental steps, and there are 20 classes to learn at each incremental step. All of the 20 initial classes are superclasses, and classes appearing in the subsequent step are subclasses of these superclasses.  Table \ref{performance_table} shows the last average pw-JS of existing methods and their combination with the proposed MTKD strategy. It is worth noting that the average improvement in IIRC-ImageNet120 is higher than that in IIRC-CIFAR100. This is due to that in our IIRC-ImageNet120, all the superclasses appear in the initial incremental step, while in IIRC-CIFAR100, superclasses may appear in subsequent steps. In IIRC-ImageNet120, the advantage of our MTKD strategy is magnified, as the superclass teacher in our MTKD strategy contains all the superclass knowledge and can teach them to the student model. Figure \ref{performance_mtkd} shows the pw-JS accuracy curve across incremental steps under the configuration of 6 incremental steps on IIRC-ImageNet120. The gap between the two accuracy curves is bigger than that in IIRC-CIFAR100, which shows our MTKD strategy is more effective when all the superclasses appearing in the initial step.

\subsection{Effectiveness of MTKD}
In this section, to figure out the reason of the great performance improvement of our MTKD strategy, we do some ablation study on the superclass teacher proposed in our MTKD. Figure \ref{confusion_mtkd} compares the confusion matrix over incremental steps with the icarl-cnn method and icarl-cnn with MTKD. The typical conventional incremental learning method icarl-cnn tends to predict the latest label (subclass label) for images since there is bright performance on the diagonal of confusion matrices, but little superclass prediction is activated. This confirms our argument: when subclasses appear, subclass knowledge may occupy their superclass knowledge effortlessly and this kind of occupation would cause the student model to rarely activate the prediction of the superclass. However, comparing the confusion matrix of icarl-cnn Figure \ref{confusion_mtkd} (d) and the matrix of icarl-cnn with MTKD after task8 Figure \ref{confusion_mtkd} (h), we can find that our MTKD strategy can help the student model activate many predictions of the superclass and maintain the discrimination of the superclass, as the activated predictions of the superclasses increase obviously (the ten columns near the leftmost). 

Figure \ref{superclass} shows the pw-JS accuracy on the superclass compared with icarl-cnn and icarl-cnn with MTKD. In Figure \ref{superclass}, we find that icarl-cnn performs poorly on the superclass after the second incremental step. Our MTKD strategy is proposed to maintain the superclass knowledge embedded in the initial model. Figure \ref{superclass} shows that the addition of our superclass teacher can help the student model improve superclass discrimination and show great improvement on superclasses, which can be regarded as contributing to the ultimate performance improvement.

% Without the knowledge from our superclass teacher, the model would rarely activate the prediction of the initial superclass. With the help of our MTKD strategy, the model can maintain the discrimination of the initial superclass.
\subsection{Comparison with methods using replay buffer of different size}
\begin{figure}[t]
\centering
\includegraphics[scale=0.45]{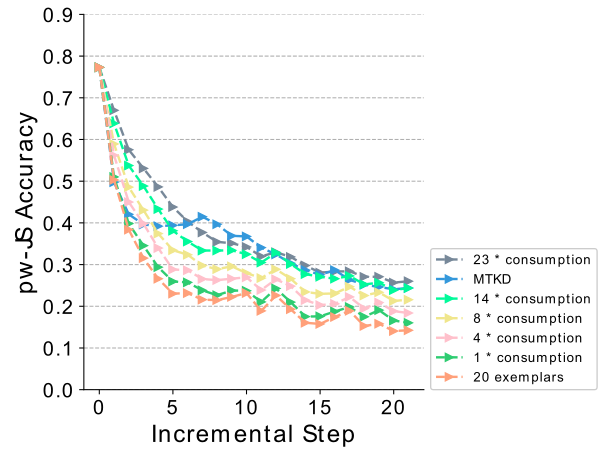}
\caption{Comparisons with MTKD and icarl-cnn using different memory consumption for rehearsal samples. 1 * consumption means 1 times the memory consumption of our superclass teacher.}
\label{explar}
\end{figure}

The additional superclass teacher preserved in our strategy would increase memory consumption. It is unfair to compare different incremental learning methods with different memory consumption. Hence, we compare our MTKD strategy with icarl-cnn using the same consumption for rehearsal samples. Our reduced Resnet-32 for IIRC-CIFAR100 would take up $1.84 MB$, which can preserve 626 samples, 5.46 samples per class for IIRC-CIFAR100. In our experiments, we preserve additional 6 images per class for the baseline with the same consumption as our method, called as 1 * consumption. To further verify the effectiveness of our strategy, we compare icarl-cnn using more memory consumption ($x$ * consumption means additional $x$ * 6 exemplars per class.) to our strategy using 20 exemplars per class. Figure \ref{explar} shows our method can achieve equivalent performance with icarl-cnn using 14 times the memory consumption of our superclass teacher (additional 84 exemplars per class). 

\subsection{Combination with the Top-k prediction restriction}
\begin{figure}[ht]
\centering
\includegraphics[scale=0.3]{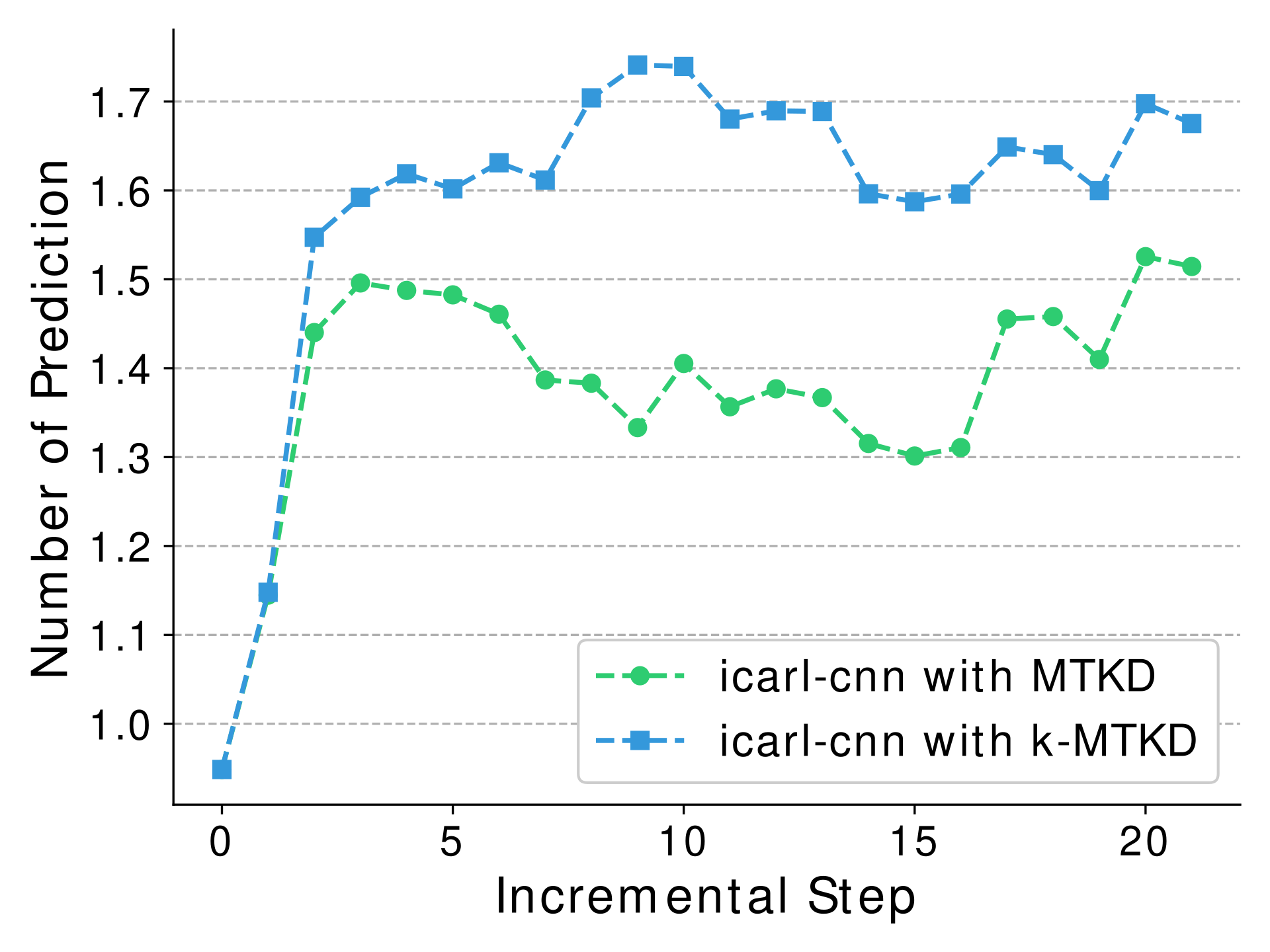}
\caption{Average number of predictions per image across the incremental steps. The Top-k prediction restriction can reduce redundant output appropriately.}
\label{prediction_num}
\end{figure}

\begin{figure}[ht]
\centering
\includegraphics[scale=0.3]{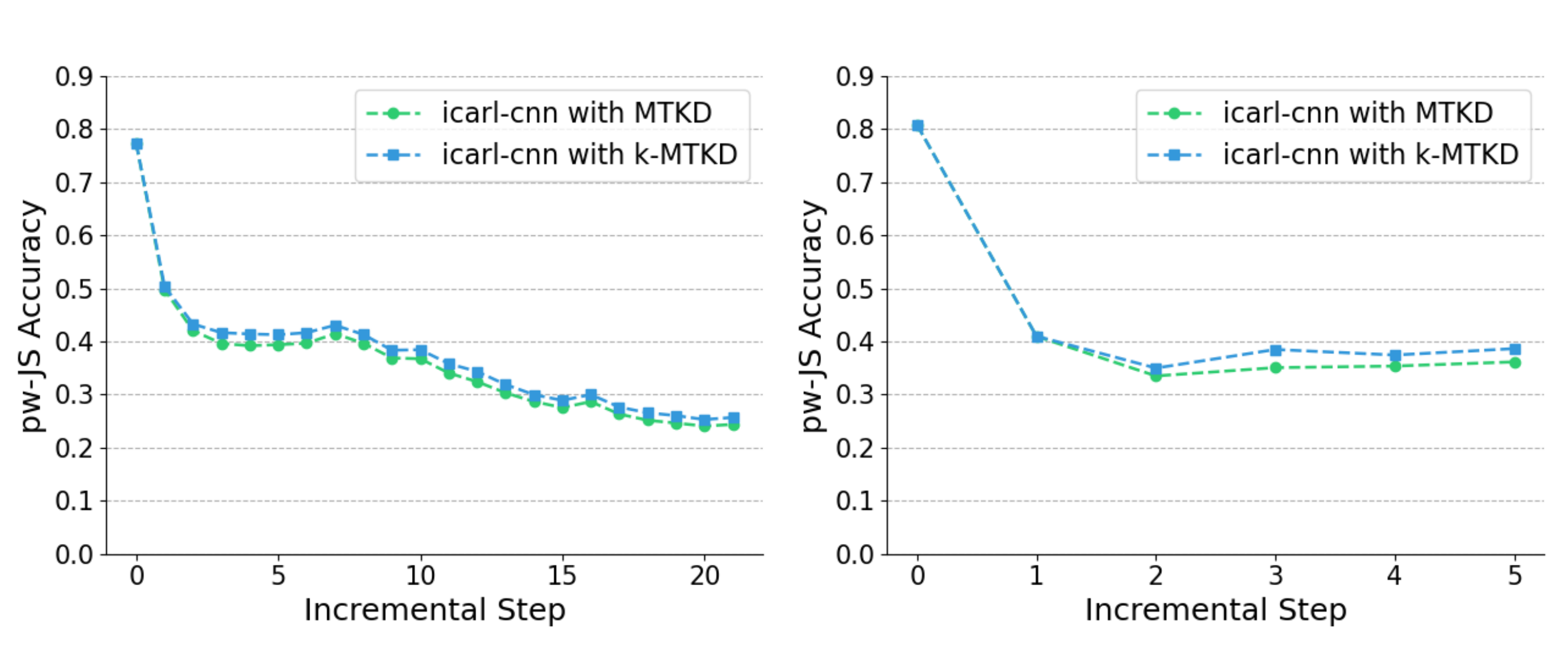}
\caption{Comparison between icarl-cnn with MTKD and that icarl-cnn with k-MTKD on IIRC-CIFAR100 (left) and IIRC-ImageNet120 (right). The Top-k prediction restriction mechanism can improve the final performance by reducing False Positive error.}
\label{topk4_18}
\end{figure}

As shown in Figure. \ref{confusion_mtkd} (g)(h), the introduction of the superclass teacher in the MTKD strategy could cause a redundant prediction problem. To conquer this issue and further improve the performance, we propose the Top-k prediction restriction mechanism combined with our MTKD strategy, called as k-MTKD.  Figure \ref{prediction_num} shows the average prediction number per image across the incremental learning steps under the baseline of icarl-cnn with MTKD and the addition of Top-k prediction restriction. It shows the Top-k prediction restriction can alleviate the prediction number per image appropriately after the $2_{th}$ incremental step. Figure \ref{topk4_18} shows that the Top-k prediction restriction can help improve performance after the $2_{th}$ incremental step, where the superclass teacher in our MTKD strategy starts at work. Table \ref{t-mtkd} shows that k-MTKD gains $11.5\%$ and $25.7\%$ performance improvement on IIRC-CIFAR100 and IIRC-ImageNet120 compared with icarl-cnn.

% \begin{table}
%     \caption{\textbf{Comparisons of average pw-JS accuracy on IIRC-CIFAR100 under 5 classes  (22 steps), 10 classes  (11 steps) and on IIRC-ImageNet120 under 20 classes  (6 steps) with our proposed MTKD strategy and state-of-the-art.}}
%     \begin{minipage}{0.98\linewidth}
%         \centering
        
%         \label{table:snetv2}
%         \resizebox{1\textwidth}{!}{
%          \begin{tabular}{l|l|l}
%           \toprule

%           \multirow{2}*{Methods} &  {CIFAR100} & ImageNet120 \\ 
%           \cline{2-3} 
%         %   \cline{4}
%           ~ & \makecell[c]{5 classes}  &\makecell[c]{10 classes} \\
%           \midrule 
%           icarl-cnn    & 14.2 & 17.6 \\ 
%             ~~~~ + MTKD  & \textbf{24.3} \textcolor{red}{\footnotesize{+10.1}} & \textbf{25.1} \textcolor{red}{\footnotesize{+7.5}} \\ 

%           \bottomrule
%         \end{tabular}
%         }
%     \end{minipage}
% \label{performance_table}
% \end{table}

\begin{table}[tbp]
 \caption{\label{t-mtkd}Comparisons of average pw-JS accuracy on IIRC-CIFAR100 under 5 classes (22 steps) and on IIRC-ImageNet120 under 20 classes (6 steps).}
 \begin{tabular}{lll}
  \toprule
  Mehtods & CIFAR100 & ImageNet120 \\
  \midrule
 icarl-cnn     & 14.2 & 12.9 \\
   ~~~~ + MTKD & 24.3 & 36.1 \\
  ~~~~ + k-MTKD& \textbf{25.7} \textcolor{red}{\footnotesize{+11.5}} & \textbf{38.6} \textcolor{red}{\footnotesize{+25.7}} \\
  \bottomrule
 \end{tabular}
\end{table}

% \begin{figure}[ht]
% \centering
% \includegraphics[scale=0.45]{topk4_18.png}
% \caption{Comparison between MTKD based icarl-cnn method and that with the Top-k prediction restriction. The Top-k prediction restriction mechanism can improve the final performance by reduce False Positive error.}
% \label{topk4_18}
% \end{figure}

% \begin{figure}
% \begin{minipage}[t]{0.5\linewidth}
% \centering
% \includegraphics[width=1.6in]{prediction_num.png}
% \caption{With the incremental step goes up, conventional incremental learning method tends to output unreasonable prediction numbers (larger than 2) per image, while with our restriction mechanism, the prediction number per image is under control.}
% \label{fig:side:a}
% \end{minipage}%
% \begin{minipage}[t]{0.5\linewidth}
% \centering
% \includegraphics[width=1.6in]{topk.png}
% \caption{The Top-k prediction restriction mechanism can improve the final performance by reduce False Positive error.}
% \label{fig:side:b}
% \end{minipage}
% \end{figure}

\section{Conclusions}
In this paper, we analyze the reasons for the poor performance of the conventional incremental learning methods in IIRC, which is that the superclass knowledge could be occupied by the knowledge of the subclass. We propose a Multi-Teacher Knowledge Distillation (MTKD) strategy to solve this issue. We use both the initial model (superclass teacher) and the last model (general teacher) to distill knowledge for our student model. In addition, using two teacher models could cause a redundant prediction problem in IIRC. We propose a simple Top-k prediction restriction mechanism combined with our MTKD strategy (k-MTKD) to reduce the unnecessary predictions. Experiments show our k-MTKD strategy can gain a great performance improvement ($11.5\%$ on IIRC-CIFAR100 and $25.7\%$ on IIRC-ImageNet120). 

% \newpage

% \begin{acks}
% \end{acks}


\bibliographystyle{ACM-Reference-Format}
% \bibliography{reference}

% \bibliography{reference}

\end{document}